\documentclass[journal]{IEEEtran}

\ifCLASSINFOpdf
  \usepackage[pdftex]{graphicx}
  \graphicspath{{../pdf/}{../jpeg/}}
  \DeclareGraphicsExtensions{.pdf,.jpeg,.png}
\else
  \usepackage[dvips]{graphicx}
  \graphicspath{{../eps/}}
  \DeclareGraphicsExtensions{.eps}
\fi

\usepackage{amsmath}
\usepackage[ruled]{algorithm2e}

\ifCLASSOPTIONcompsoc
 \usepackage[caption=false,font=normalsize,labelfont=sf,textfont=sf]{subfig}
\else
 \usepackage[caption=false,font=footnotesize]{subfig}
\fi

\usepackage{amsmath}
\usepackage[numbers, square]{natbib}
\usepackage{booktabs}
\usepackage{hyperref}
\hypersetup{
colorlinks=true,
linkcolor=black,
anchorcolor=black,
citecolor=green}
\usepackage{bbm}
\usepackage{tabularx}
\usepackage[table,xcdraw]{xcolor}
\usepackage{xparse}
\usepackage{multirow}
\usepackage{adjustbox}

\begin{document}

\title{Long and Short-Term Constraints Driven Safe Reinforcement Learning for Autonomous Driving}

\author{
        Xuemin Hu,
        Pan Chen,
        Yijun Wen,
        Bo Tang,~\IEEEmembership{Senior Member, IEEE},
        and Long Chen,~\IEEEmembership{Senior Member,~IEEE}
\thanks{This work was supported in part by the National Natural Science Foundation
of China under Grants 62273135 and 62373356, and in part by the Postgraduate Education and Teaching Reform Research Project of Hubei University under
Grant 1190017755. (Corresponding author: Long Chen)}

\thanks{Xuemin Hu, Pan Chen, and Yijun Wen are with the School of Artificial Intelligence, Hubei University, Wuhan 430062, China, and also with the Key Laboratory of Intelligent Sensing System and Security (Hubei University), Ministry of Education, Wuhan, Hubei, 430062, China. (e-mail: huxuemin2012@hubu.edu.cn; 178234395@qq.com; wenjunazi@gmail.com)}

\thanks{Bo Tang is with the Department of Electrical and Computer Engineering, Worcester Polytechnic Institute, Worcester, MA, 01609, USA. (e-mail: btang1@wpi.edu)}

\thanks{Long Chen is with the State Key Laboratory of Multimodal Artificial Intelligence Systems and the State Key Laboratory of Management and Control for Complex Systems, Chinese Academy of Sciences, Beijing, 100190, China. Long Chen is also with WAYTOUS Inc., Beijing, 100083, China, the Guangdong Laboratory of Artificial Intelligence and Digital Economy (SZ), Shenzhen 518107, China, and the Institute of Artificial Intelligence and Robotics, Xi'an Jiaotong University, Xi'an, 710049, China (e-mail: long.chen@ia.ac.cn).}
}


\mark{T This work has been submitted to the IEEE for possible publication. Copyright maybe transferred without notice, after which this version may no longer be accessible.}

\maketitle

\begin{abstract}
Reinforcement learning (RL) has been widely used in decision-making and control tasks, but the risk is very high for the agent in the training process due to the requirements of interaction with the environment, which seriously limits its industrial applications such as autonomous driving systems. Safe RL methods are developed to handle this issue by constraining the expected safety violation costs as a training objective, but the occurring probability of an unsafe state is still high, which is unacceptable in autonomous driving tasks. Moreover, these methods are difficult to achieve a balance between the cost and return expectations, which leads to learning performance degradation for the algorithms. In this paper, we propose a novel algorithm based on the long and short-term constraints (LSTC) for safe RL. The short-term constraint aims to enhance the short-term state safety that the vehicle explores, while the long-term constraint enhances the overall safety of the vehicle throughout the decision-making process, both of which are jointly used to enhance the vehicle safety in the training process. In addition, we develop a safe RL method with dual-constraint optimization based on the Lagrange multiplier to optimize the training process for end-to-end autonomous driving. Comprehensive experiments were conducted on the MetaDrive simulator. Experimental results demonstrate that the proposed method achieves higher safety in continuous state and action tasks, and exhibits higher exploration performance in long-distance decision-making tasks compared with state-of-the-art methods.
\end{abstract}


\IEEEpeerreviewmaketitle

\section{Introduction}
Autonomous driving (AD) is rapidly developing, and great achievements have been made in recent years \cite{10440197, 10399367, Hu2024Howl, chen2023milestones, liang2023robust}. As a new technology, end-to-end autonomous driving, which learns driving policies from collected data or environmental interaction rather than pre-defined complex rules, attracted researchers' attention \cite{teng2023motion, hu2020learning, hu2023learning}. In comparison to traditional pipeline-based methods \cite{schnelle2016driver, hu2018dynamic}, it simplifies the architecture of AD systems and replaces the multi-module system with a single end-to-end model, where raw sensor data are directly input to generate corresponding driving commands such as the steering angle, braking, and throttle.

In recent years, reinforcement learning (RL) has become the mainstream framework for end-to-end autonomous driving \cite{chen2020conditional} since it does not require a large number of labeled samples like imitation learning. The vehicle agents can be trained by interacting with the environment without human intervention and adjust their behaviors based on the feedback provided by the environment. However, this trial-and-error property brings potential safety risks for training an effective model in a practical end-to-end AD system.

\begin{figure}[!t]
\centering
\includegraphics[width=3.0in]{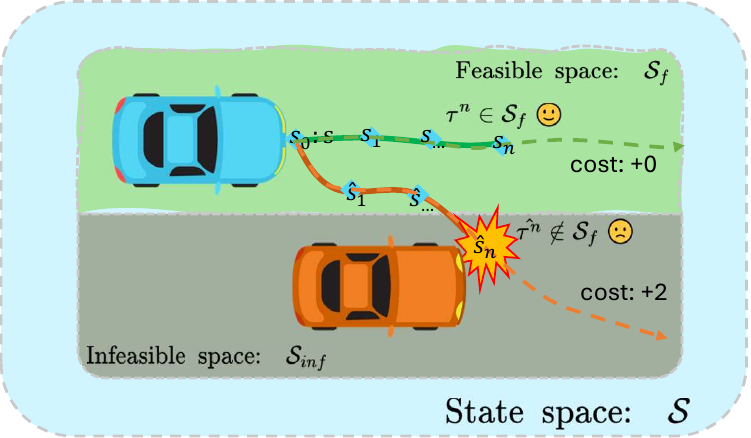}
\DeclareGraphicsExtensions.
\caption{Long and short-term constraints for AD and the state space definitions. The reliable state space $\mathcal{S}_{f}$ indicates the state set where no danger occurs. The infeasible state space $\mathcal{S}_{inf}$ indicates the state set where a danger is imminent or already present. The short solid and long dotted lines denote the short and long trajectories of the vehicle, respectively. The blue dots represents the state in the short trajectories.}
\label{fig_1}
\end{figure}

Safe reinforcement learning, which is often modeled as a constrained Markov decision process (CMDP) \cite{altman2021constrained}, is developed to handle the safety issue of traditional RL methods by introducing a constraint in the training process. 
In these methods, a cost is usually predefined to evaluate the risk of the agent at current time step.
The generalized constraint emphasizes that the expected cost should be less than a predetermined threshold. Many safe RL methods \cite{chow2017risk, achiam2017constrained, tessler2018reward, yang2020projection, zhang2020first, stooke2020responsive} use the expected cost, similar to the expected reward, as a secondary objective which mainly includes the descriptions of decision risks or losses, which is a long-term objective constraint. 
These methods exhibit good safety performance, convergence, and stability of network training in many intelligent control tasks. However, they only guarantee the safety after the training convergence, but do not consider the state safety of the agent in training process, which may lead to serious accidents when applying in autonomous driving. Moreover, using the expected cost as a constraint is a modification of the goal-oriented learning approach that is similar to reward shaping, which may lead to inconsistency in the optimal strategy and result in performance reduction.

In this paper, we propose a novel safe RL method with Long and Short-Term Constraints (LSTC) for decision-making tasks in end-to-end autonomous driving, as shown in Fig.\ref{fig_1}. The long-term constraint is designed based on the cost signals, with the goal of keeping the vehicle safe throughout a complete task. The short-term constraint, where a learnable forward-domain safety model checker that is designed to verify the state safety of a trajectory, aims to enhance the short-term state safety that the vehicle explores. To solve the optimization problem for the proposed LSTC, we also develop a safe RL method with dual-constraint optimization based on the Lagrange multiplier. In conclusion, the main contributions are as follows.
\begin{itemize}
 \item We propose the long and short-term constraints for safe RL, which aims to address the issue of unrestricted exploration in existing RL-based methods and enhance the agent safety in the training process.
 \item To improve safety performance without sacrificing learning performance, we develop a safe RL method with dual-constraint optimization for autonomous driving, which can solve the constrained optimization problem of the proposed LSTC. 
 \item Comprehensive experiments were conducted on the MetaDrive simulator, and experimental results demonstrate that the proposed method outperform state-of-the-art (SOTA) methods in term of the success rate of driving and robustness in complex driving scenarios.
\end{itemize}

\section{RELATED WORK}

\subsection{End-to-end Autonomous Driving}
Traditional pipeline-based autonomous driving has achieved remarkable success in recent years \cite{OS, SOS}, but it remains some limitations including the difficulty of high computational costs and cumulative errors between modules \cite{jiang2022vehicle}. End-to-end methods can handle these issues by replacing the complex modules with a single deep neural network model \cite{chen2022Improved}, and they are often categorized into two types: imitation learning (IL) and reinforcement learning (RL) \cite{teng2023motion}.

IL-based methods train an agent to learn the optimal strategy by imitating the expert demonstration. One of the most important and widely used methods is the behavior cloning (BC) \cite{bojarski2016end}. Codevilla et al. \cite{codevilla2018end} proposed the Conditional Imitation Learning (CIL) model, which inputs monocular images, self-vehicle speed measurements, high-level driving decision-making and directly predicts the vehicle's lateral and longitudinal controls. Many researchers expand on \cite{codevilla2018end} by incorporating additional input like global routes, positional data, or point clouds \cite{wang2019end, peng2020imitative, teng2022hierarchical}. Due to the richer input data, these methods exhibit impressive adaptability and resilience across a wide array of scenarios.
BC demonstrates proficiency when dealing with states within the training distribution. However, the generalization ability to new states diminishes due to the cumulative errors in actions, which limits its industrial applications. In addition, IL-based methods require a large number of labeled data for training, on which many approaches \cite{zhang2016query, codevilla2018end, chen2019deep} rely to achieve good performance. However, it's challenging to make the collected data cover all the possible states in complex urban driving scenarios.

RL-based methods \cite{huang2017parameterized,xu2018reinforcement} gradually attract researchers' attention in recent years due to the advantage of not requiring a large number of labeled data. The agent in RL learns to make decisions by trial and error in the process of interacting with the environment, in which it receives the feedback in the form of rewards or penalties for its actions and aims to maximize the total cumulative reward. Alizadeh et al. \cite{alizadeh2019automated} train a agent with the deep Q network (DQN) to make high-level decisions.
In real-world scenarios, Kendall et al. \cite{lillicrap2015continuous} use the deep deterministic policy gradient (DDPG) algorithm, and the vehicle can drive in simple environments like human drivers.
\cite{saxena2020driving, ye2020automated, guan2020centralized, wu2021deep} indicate that the Proximal Policy Optimization (PPO) algorithm performs well in the tasks with continuous driving actions. 
However, these methods are developed in simulated environments since they ignore the safety problem in the training and testing processes.

Safety consideration is increasingly being emphasized, especially for AD tasks in the real environment.
Introducing safety checker is a common approach, which can recognize whether an action belongs to the set of safe actions.
Zhang et al. \cite{zhang2016query} propose SafeDAgger, where the safety policy can predict whether the decision made by the primary policy is safe. If the decision is predicted to be unsafe, the reference policy is activated to take over. 
The reference policy serves as the final safeguard to determine the safety of the ultimate decision. 
Bouton et al. \cite{bouton2019reinforcement} design a probability model to predict the probability of each decision and compare this probability with a threshold to identify safe actions, enabling the vehicle to safely reach the destination. 
The probability model exhibits high accuracy in discrete state-action spaces, but it is not applicable in continuous high-dimensional spaces. 
Inspired by the idea of safety checker, we focus on the long and short-term safety of decisions and propose a safe RL method for continuous state-action spaces in AD, which can validate the safety of forward state trajectories of the vehicle agent.

\subsection{Safe reinforcement learning}
Safe reinforcement learning is commonly modeled as the constrained Markov decision process \cite{altman2021constrained}, which extends the constraint concept to the Markov decision process (MDP). Generally, a constraint is defined as the discounted expected cost that is less than a threshold. The discounted expected cost, similar to a value function, is usually calculated through a cost value function.

The Lagrange method is widely used to solve the CMDP problem which is transformed into an unconstrained dual problem by implicitly penalizing the original objective function \cite{ray2019benchmarking}. 
Tessler et al. \cite{tessler2018reward} introduce a constrained actor-critic method by incorporating a constraint as the penalty signal into the reward function. Roy et al. \cite{roy2021direct} introduce constraints based on the frequency of behavior occurrences in SAC (Soft Actor-Critic) and utilize the Lagrange method to automatically balance each behavioral constraint.

Another kind of approaches involve the trust region method, which tackles constrained policy optimization problems by modifying the policy gradient of the trust region. These methods project the policy onto a safe feasible set during each iteration, ensuring that the policy remains within the expected constraint boundaries.
Achiam et al. \cite{achiam2017constrained} propose a constrained policy optimization (CPO) method which constructs an auxiliary function to approximate the original objective function, uses a second-order Taylor expansion to approximate the objective and cost functions, and employs backtracking line search to update policies to satisfy constraints.
Compared to \cite{achiam2017constrained}, Yang et al. utilize the projected gradient descent method, employing projection onto the constraint set to ensure compliance with the constraints.
Zhang et al. \cite{zhang2022penalized} design a penalty function based on PPO \cite{schulman2017proximal} to transform the constrained cost into an unconstrained problem.

The aforementioned methods provide solutions for constrained optimization problems and have been proven to be effective. However, they overlook the state safety of the agent during the training process, essentially requiring trial and error to minimize errors. In contrast, our approach enhances the agent safety during both training and deployment through a dual-constraint mechanism.

\section{Preliminaries and problem formulation}
Following the CMDP theory, we expand the state-based constraint and provide related definitions for the LSTC method.
\newtheorem{definition}{Def.}
\begin{definition}[\textbf{Feasible state and infeasible state}]\label{definition1}
    In our method, the whole state space of the vehicle is divided into two parts: feasible state space $\mathcal{S}_{f}$ and infeasible state space $\mathcal{S}_{inf}$, as shown in Fig. \ref{fig_1}. A feasible state $s_{f}\in\mathcal{S}_{f}$ indicates that there is no risk associated with this state, while the infeasible state $s_{inf}\in\mathcal{S}_{inf}$ suggests that a vehicle in this state may encounter collisions with other vehicles, obstacles, buildings, etc. We differentiate them through specific rules, including speed limits, time-to-collision (TTC), and heading angle deviation, etc. following \cite{yang2023model}. 
\end{definition}

\begin{definition}[\textbf{State trajectory}]\label{definition2}
The state trajectory $\tau^{n}$ with total $n$ time steps, which is generated from state $s_{t}$ following policy $\pi$, is defined as Eq.\ref{eq1}.
\begin{equation}\label{eq1}
    \begin{split}
        \tau^{n}&=\left\{s_{t}|\pi\right\}_{t}^{t+n}  
                =\left\{s_{t}, s_{t+1}, s_{t+2}, s_{t+3},..., s_{t+n}|\pi\right\},
    \end{split}
\end{equation}
\end{definition}
where the subscript denotes the index of the time step and $t$ represents the current time step. The objective of safe RL is defined to maximize the expected reward while satisfying the safety constraints. Specifically, we establish the stringent safety constraint for the AD task and define the sets of the feasible policy and objective policy.

\begin{definition}[\textbf{Feasible policy and objective policy}]\label{definition3}
If the policy $\pi$ ensures that all the state trajectories from state $s$ to the task termination remain in $\mathcal{S}_{f}$, we refer it as a feasible policy, and the set composed of all feasible policies is referred to as the feasible policy set $\Pi_{c}$, which can be formulated as Eq. \ref{eq3}.
\begin{equation}\label{eq3}
    \begin{split}
        \Pi_{c}=\left\{\pi | s_t\in \mathcal{S}_{f}, \forall t \in\{0,1,2,3,\cdots\}\right\}.
    \end{split}
\end{equation}
The objective policy is described as $\pi^*$, as shown by Eq. \ref{eq4}, which is used to maximize the expected reward from $\Pi_c$.
\begin{equation}\label{eq4}
    \begin{split}
        \pi^{*}=\arg \max _{\pi \in \Pi_{c}}J_{\pi}(\theta),
    \end{split}
\end{equation}
\end{definition}
where $\theta$ indicates the parameters in policy $\pi$, and 
\begin{equation}\label{eq4_2}
    \begin{split}
        J_{\pi}(\theta)&=\mathbbm{E}_{s\sim\rho^{\pi}(s)}\left\{V_{\pi}(s)\right\},\\
        V_{\pi}(s) &=\mathbbm{E}_{\pi}\left[\sum_{t=0}^{\infty} \gamma^{t} r_{t+1} \mid s_{0}=s\right],
    \end{split}
\end{equation} 
where $J_{\pi}(\theta)$ denotes the objective function. $V_{\pi}(s)$ is the state value function. $\gamma$ is the discount factor. $s_0$ is the initial state, and $r$ is the one-step reward from the environment. $\rho^{\pi}(s)$ denotes the distribution of state $s$ based on the policy $\pi$.


\section{Methodology}
CMDP-based methods use the cost value function as the constraint, which lacks state-based safety and is potentially dangerous for AD vehicles. In this paper, we aim for the vehicle to drive within a reliable region during both exploration and deployment. For this purpose, we extend the state-based constraint by using forward state trajectories for the safety verification and propose our safe RL method with LSTC in order to constrain the vehicle's exploration in the safe space and enhance the vehicle safety during the training process.

\subsection{Long and short-term constraints}
RL methods typically maximize the expected reward to optimize the policy. In this case, the agent learns its policy by trial and error without considering risks to pursue a great reward in traditional RL methods. In this paper, we propose the LSTC method to fills the safety gap.

\textbf{Long-term constraint:}  
In RL methods, the objective function aims to maximize long-term expected returns which can be calculated by the reward, as shown in Eq. \ref{eq4_2}. Similar to the the expected return, the goal of the long-term constraint is to minimize the expected costs for the agent safety in safe RL methods. Inspired by the idea of maximizing the expected return in RL methods, we design the long-term constraint following traditional CMDP-based safe RL methods \cite{altman2021constrained} to reduce the long-term risk of the agent. The long-term constraint is described by Eq. \ref{c1}.
\begin{equation}\label{c1}
    \begin{split}
        C_{\pi}(\theta) < b,
    \end{split}
\end{equation}
where $C_{\pi}(\theta)$ is the expected cost function, and $b$ is the threshold. Similar to the objective function and state value function shown by Eq. \ref{eq4_2}, we define the expected cost function $C_{\pi}(\theta)$ and cost value function $V_{\pi}^{c}(s)$ by Eq. \ref{eq5}.
\begin{equation}\label{eq5}
    \begin{split}
        C_{\pi}(\theta) &= \mathbbm{E}_{s\sim\rho^{\pi}(s)}\left\{V_{\pi}^{c}(s)\right\},\\
        V_{\pi}^{c}(s) &=\mathbbm{E}_{\pi}\left[\sum_{t=0}^{\infty} \gamma^{t} c_{t+1} \mid s_{0}=s\right],
    \end{split}
\end{equation}
where $c_{t+1}$ is the cost value from the environment at the time step $t+1$, which is calculated by the predefined cost function like the reward function in RL methods. The details of calculating the reward and cost at a time step will be shown in Sec. V-A. The cost value function $V_{\pi}^{c}(s)$ is used to estimate the long-term cost that the agent can expect when executing a certain policy in different states, so we refer to this constraint as the long-term constraint. Since the proposed long-term constraint is designed to modify the objective function so as to decrease the final cost value, it can enhance the agent safety from the perspective of the long-term learning objective.

\textbf{Short-term constraint:} AD vehicles need to consistently drive within the feasible state space $\mathcal{S}_f$ for safety. Although the proposed long-term constraint can reduce the risk after training convergence, it struggles to handle the tasks of requiring the state safety such as autonomous driving. Nevertheless, as defined in Def. \ref{definition3}, our objective policy is to select the optimal policy within the feasible space, where a feasible policy can ensure that AD vehicles are safe in every state under that policy since the safety in each state is particularly crucial for AD vehicles.

In order to address this issue, we decompose the driving process from the initialization to termination into sequential states, and expect each state along with several subsequent states in the vehicle's trajectory to fall in the feasible state space, which aims to ensure that the AD vehicle always navigates in the feasible state space and meets safety requirements, as illustrated in Fig. \ref{fig_1}. In other words, each $n$-step state trajectory $\tau^{n}$ starting from the state $s$ needs to adhere to a feasible policy $\pi$ to remain in the feasible space $\mathcal{S}_{f}$. According to this idea, we develop the short-term constraint based on each state safety of the AD vehicle, as formulated in Eq. \ref{6}.
\begin{equation}\label{6}
    \begin{split}
        \tau^{n}=\left\{s_{t}|\pi\right\}_{t}^{t+n} \in \mathcal{S}_{f}.
    \end{split}
\end{equation}

\begin{figure}[tb!]
    \centering
    \includegraphics[width=0.9\linewidth]{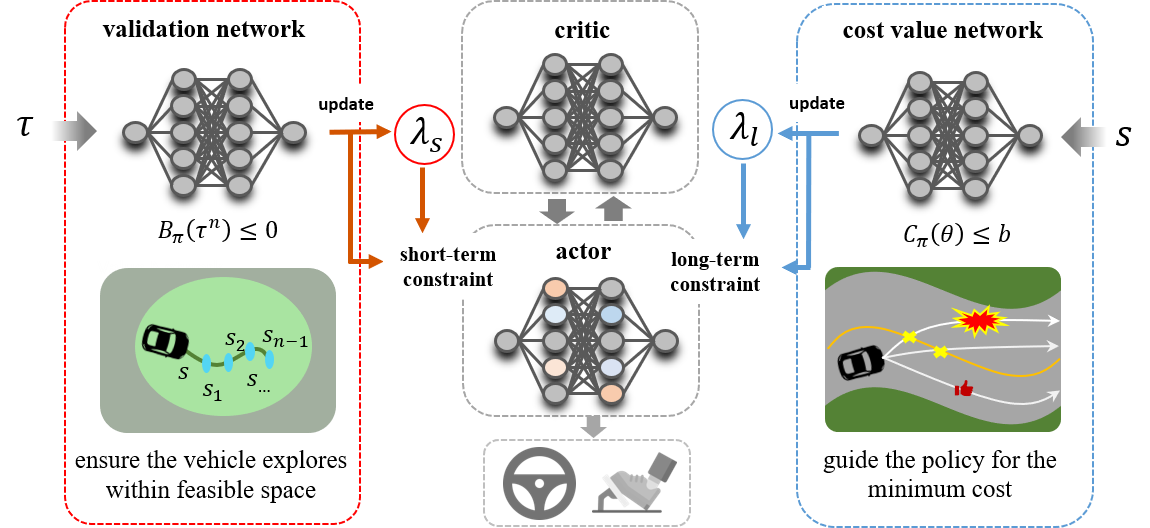}
    \caption{Safe RL method with dual-constraint optimization. The actor-critic framework includes the cost value network and validation network. The cost value network takes the current state $s$ as the input and evaluates the expected cost. The validation network takes a state trajectory $\tau$ as the input and validates the short-term feasibility of decisions. When the constraints are not satisfied, the Lagrange multipliers are updated. The Lagrange multipliers act as a penalty in participating in policy updating, in order to learn a safe policy.}
    \label{net}
\end{figure}

To satisfy the constraint in Eq. \ref{6}, we define the validation function $B_{\pi}(\tau^{n})$, which is represented by the validation network in Fig. \ref{net}, and map this relationship by Eq. \ref{8_2}.
\begin{equation}\label{8_2}
\left\{
\begin{array}{l}
    B_{\pi}(\tau^{n}) \leq 0  \quad if   \quad \tau^n \in \mathcal{S}_{f},   \\
    B_{\pi}(\tau^{n}) > 0  \quad  others,
\end{array}
\right.
\end{equation}
which means that the current action can generate a safe policy at the next $n$ time steps, and vice versa.


\textbf{Objective function with LSTC:} After combining the long and short-term constrains, we design the objective function by Eq. \ref{8}.
\begin{equation}\label{8}
    \begin{split}
        \underset{\pi}{\max}J_{\pi}(\theta) =&\mathbbm{E}_{s\sim\rho^{\pi}(s)}\left\{V_{\pi}(s)\right\},\\ 
        s.t.\quad &C_{\pi}(\theta) \leq b,   \quad 
         B_{\pi}(\tau^{n}) \leq 0.
    \end{split}
\end{equation}

Inspired by the primal-dual gradient descent \cite{paternain2022safe}, the dual ascent updating is designed to converge towards the saddle point of the maximization problem. Therefore, we propose a Lagrange function $L(\theta,\lambda_s, \lambda_l)$ by introducing two Lagrange multipliers, as shown by Eq. \ref{9}.
\begin{equation} \label{9}
    \begin{split}
        &\underset{\pi}{\max}\underset{\lambda_s, \lambda_l\geq0}{\min}L(\theta,\lambda_s, \lambda_l)\\
        &=J_{\pi}(\theta)-\lambda_l(C_{\pi}(\theta) - b)-\lambda_s B_{\pi}(\tau^{n}),
    \end{split}
\end{equation}
where $\lambda_s$ and $\lambda_l$ are the Lagrange multipliers of the long and short constraints, respectively.
The solution of the Lagrange function defined in Eq. \ref{9} could be solved by alternatively updating the policy and the two Lagrange multipliers, as shown by Eq. \ref{10}.

\begin{equation}\label{10}
\left\{
   \begin{array}{lr}
        \theta \gets \theta + \alpha_{\theta}\nabla_{\theta} L(\theta,\lambda_s, \lambda_l),  &\\
        \lambda_s \gets \lambda_s - \alpha_s \nabla_{\lambda_s} L(\theta,\lambda_s, \lambda_l), &\\
        \lambda_l \gets \lambda_l - \alpha_l \nabla_{\lambda_l} L(\theta,\lambda_s, \lambda_l),    &
   \end{array}  
\right.
\end{equation}
where $\alpha_{\theta}$, $\alpha_s$, $\alpha_l$ represent the step size for gradient descent or ascent. $\lambda_s$ and $\lambda_l$ are essentially penalty coefficients that increase the penalty intensity when the constraints are violated. When the short-term constraint shown by Eq. \ref{8_2} is not satisfied, $\lambda_s$ will strongly increase the penalty to Lagrange function to ensure forward area feasible. Similarly, $\lambda_s$ can enhance the safety throughout the entire task by increasing the penalty when the long-term constraint shown by Eq. \ref{c1} is not satisfied. In this case, the joint constraint from both short and long terms enables the vehicle agent for safe exploration.

\subsection{Safe RL method with dual-constraint optimization}
In order to apply the proposed LSTC in autonomous driving, we develop a safe RL method with dual-constraint optimization based on the Actor-Critic (AC) framework, as shown in Fig. \ref{net}. The proposed safe RL method contains the validation network, the cost value network as well as the actor and critic networks in the AC framework. The policy (actor) network takes the state $s$ as the input and outputs the vehicle controls including the steering angle and accelerator value for the vehicle. The value (critic) network takes the state $s$ as the input and outputs the expected reward. The validation network is designed to approximate the validation function $B(\tau^{n})$. In our method, it serves the short-term constraint and used to assess the feasibility of the state trajectory, keeping the short-term state-based safety for the vehicle. The cost value network, serving the long-term constraint, is used to estimate the expected cost of the current state and provide the long-term safety. 

To represent these networks in our safe RL method, we transform the solution of the Lagrange function in Eq. \ref{9} into the issue of parameter optimization of these networks including the policy network $\pi_{\theta}$ parameterized by $\theta$, the value network $V_{\omega_{r}}$ parameterized by $\omega_{r}$, the validation network $B_{\phi}$ parameterized by $\phi$, and the cost value network $V^{c}_{\omega_c}$ parameterized by $\omega_{c}$, as well as the long and short-term Lagrange multipliers  $\lambda_l$ and $\lambda_s$.

The objective in the proposed LSTC-based safe RL method, as shown by Eq. \ref{8}, is to find a policy that maximizes the reward under the long and short-term constraints. To sovle this problem, we use PPO \cite{schulman2017proximal} as our RL learner, whose objective function is show by Eq. \ref{14}.
\begin{equation}\label{14}
    \begin{split}
        J_{\pi}(\theta) = \hat{\mathbbm{E}}_{t}[\min(r_{t}(\theta) \hat{A}_{t}, \operatorname{clip}(r_{t}(\theta), 1-\epsilon, 1+\epsilon) \hat{A}_{t})],
    \end{split}
\end{equation}
where  $r_{t}(\theta)=\frac{\pi_{\theta }(a|s)}{\pi_{old}(a|s)}$, and $\hat{\mathbbm{E}}_{t}$ denotes the empirical expectation over time steps. $\hat{A}_{t}$ is the advantage function, which is shown by Eq. \ref{15}.
\begin{equation}\label{15}
    \hat{A}_{t} = Q_{\pi}(s, a) - V_{\pi}(s),
\end{equation}
where $Q_{\pi}(s, a)$ is the action value function. 

The advantage function plays a pivotal role by quantifying the advantage of taking a specific action in a given state over the average action, and it can be estimated using generalized advantage estimation (GAE) \cite{schulman2015high}.
The short-term constraint in the Lagrange function is calculated using the data collected from the old policy $\pi_{old}$, hence we also apply the importance sampling to weight $ B_{\pi}(\tau^{n})$, as shown by Eq. \ref{16}.
\begin{equation}\label{16}
    \begin{split}
        \underset{\pi}{\max}&\underset{\lambda_s, \lambda_l\geq0}{\min} L(\theta,\lambda_s, \lambda_l)\\
        &=J_{\pi}(\theta)-\lambda_l(C_{\pi}(\theta) - b)-\lambda_sr_{t}(\theta)B_{\pi}(\tau^{n}) \\
        & = \hat{\mathbbm{E}}_{t}[\min(r_{t}(\theta) (\hat{A}_{t}-\lambda_l \hat{A}_{t}^{c} - \lambda_s B_{\pi}(\tau^n)), \\
        &\operatorname{clip}(r_{t}(\theta), 1-\epsilon, 1+\epsilon) (\hat{A}_{t}-\lambda_l \hat{A}_{t}^{c} - \lambda_s B_{\pi}(\tau^n)))].
    \end{split}
\end{equation}

The term $(\hat{A}_{t}-\lambda_l \hat{A}_{t}^{c} - \lambda_s B_{\pi}(\tau^n))$ determines the orientation and magnitude of policy updating, which is influenced by the advantage function, cost advantage function, and validation function. The Lagrange multipliers  $\lambda_l$ and $\lambda_s$ serve as penalty factors for regulating the penalty intensity.

The value network and cost value network optimize the parameters through minimizing the temporal-difference (TD) error, thereby improving the estimation of cumulative reward and cost.
We use the mean squared error (MSE) as their loss functions, as shown by Eqs. \ref{17} and \ref{18}, respectively.
\begin{equation}\label{17}
    \begin{split}
        L_{C_{r}}(\omega_r) = \mathbbm{E}[(r_{t+1}+\gamma V_{\omega_r}(s_{t+1})-V_{\omega_r}(s))^2],
    \end{split}
\end{equation}
\begin{equation}\label{18}
    \begin{split}
        L_{C_{c}}(\omega_c) = \mathbbm{E}[(c_{t+1}+\gamma V_{\omega_c}(s_{t+1})-V_{\omega_c}(s))^2],
    \end{split}
\end{equation}
where $L_{C_{r}}(\omega_r)$ and $ L_{C_{c}}(\omega_c)$ are the loss functions of the value network and the cost value network, respectively. 
$r_{t+1}+\gamma V_{\omega_r}(s_{t+1})$ and $c_{t+1}+\gamma V_{\omega_c}(s_{t+1})$ are the TD targets. $V_{\omega_r}(s)$ and $V_{\omega_c}(s)$ are the estimations of the value function and the cost value function, respectively. 

\begin{algorithm}[t]
\caption{Safe RL with dual-constraint optimization}\label{algo1}
Initialize network parameters: policy network $\theta$, value network $\omega_r$, cost-value network $\omega_c$, validation network $\phi$, long and shot-term lagrange multipliers $\lambda_l$, $\lambda_s$, and learning rates $\alpha_{\theta}$, $\alpha_{\omega_r}$, $\alpha_{\omega_c}$, $\alpha_{\phi}$, $\alpha_{\lambda_s}$, $\alpha_{\lambda_l}$.\\
\For{$epoch$}{
    \textit{\# Sampling} \\ 
    \For{Episode}{
        initialize the sampling pool $\mathcal{D}$ and initial state $s_{0}$.\\
        \For{step $t$}{
             Sample from environment and store $(s,a,r)$ in $\mathcal{D}$
        }
    }
    \textit{\# Optimization} \\ 
    Sample minibatch from $\mathcal{D}$,
    First update the Lagrange Multiplier,\\
    \If{short-term constraint violation}{
         $\lambda_s \gets \lambda_s + \alpha_{\lambda_s}\nabla_{\lambda_s}L(\theta,\lambda_s, \lambda_l)$        
    }
    \If{short-term constraint violation}{
         $\lambda_l \gets \lambda_l +   \alpha_{\lambda_l}\nabla_{\lambda}L(\theta,\lambda_s, \lambda_l)$
    }
    Update the validation network by Eq. \ref{13}:
    $\phi \gets \phi + \alpha_{\phi}\nabla_{\phi}L(\phi)$\\
    Update the policy network by Eq. \ref{14}:
    $\theta \gets \theta + \alpha_{\theta}\nabla_{\theta}L(\theta, \lambda, \phi)$\\
    Update the value network by Eq. \ref{17}:
    $\omega_r \gets \omega_r + \alpha_{\omega_r}\nabla_{\omega}L_{\omega_r}(\omega_r)$\\
    Update the cost-value network Eq. \ref{18}:
    $\omega_c \gets \omega_c + \alpha_{\omega_c}\nabla_{\omega}L_{\omega_c}(\omega_c)$
}
\end{algorithm}

The validation network can assess whether the current decision will keep the vehicle safe in the short-term future, and it is updated using the sampled data in the dataset. The optimization goal is to minimize the error of validating the next $n$-step trajectory.

Based on the property of $B(\tau^{n})$, we design the loss function according to the state violation. If all states in the $n$-step state trajectory are in the feasible state space $\mathcal{S}_{f}$, we design a loss function $L_{f}(\phi)$ , as shown in Eq. \ref{11}.
\begin{equation}\label{11}
    \begin{split}
        L_{f}(\phi) =  \mathbbm{E}_{\tau^{n}} [\max(B_{\phi}(\tau^{n}),0)], \forall s \in \mathcal{S}_{f}, s \in \tau^{n}. \\
    \end{split}
\end{equation}

Otherwise, if there is a state within the infeasible state space $\mathcal{S}_{inf}$ in the trajectory, the loss function is designed by Eq. \ref{12}.
\begin{equation}\label{12}
    \begin{split}
        L_{inf}(\phi) =  \mathbbm{E}_{\tau^{n}} [\max(-B_{\phi}(\tau^{n}),0)], \exists s \in \mathcal{S}_{inf}, s \in \tau^{n}.\\
    \end{split}
\end{equation}

The loss function of the validation network is shown by Eq. \ref{13}.
\begin{equation}\label{13}
    \begin{split}
        L_{B}(\phi) = L_{f}(\phi)+L_{inf}(\phi).
    \end{split}
\end{equation}

The pseudocode of the safe RL with dual-constraint optimization is shown in Alg. \ref{algo1}.

\section{EXPERIMENTS}
\subsection{Experiment Settings}
The MetaDrive \cite{li2022metadrive} simulator is used as the training and evaluation environment in this paper since it has realistic vehicle dynamics and can provide various road maps which are randomly composed from elementary components such as the straight road, curve, intersection, roundabout, ramp, and fork. Each road map has a start point and a destination, connected with a route in the form of a series of navigation points. It is the most popular benchmark in recent studies about safe RL in autonomous driving such as \cite{zhang2022saferl, peng2022safe}. In our experiments, the AD vehicle's task is to navigate from the start point to the destination based on navigation points.

A 49-dimensional vector is defined as the observation for the RL learner, including the following categories of information:
\begin{itemize}
    \item A 30-dimensional vector denoting the distance detected by a 2D-lidar to other objects with $50m$ maximum detecting distance centering at the ego vehicle.
    \item A 9-dimensional vector summarizing the ego vehicle’s state such as the steering, heading, velocity, and relative distance to the left and right boundaries.
    \item A 10-dimensional vector representing the distance from the ego vehicle to checkpoints along the road. These checkpoints are evenly distributed along the route at intervals of $50m$.
\end{itemize}

The reward function, as shown by Eq. \ref{rewardF}, includes the dense rewards based on speed, driving distance, yaw rate, and steering angle, as well as the sparse reward for departing from lane and reaching the destination. 
\begin{equation}\label{rewardF}
    \begin{aligned}
        R=&{{c}_{1}}{{R}_{dis}}+{{c}_{2}}{{R}_{speed}}+{{c}_{3}}{{R}_{yaw}} 
        +{{c}_{4}}{{R}_{steering}}+{{R}_{term}},
\end{aligned}
\end{equation}
where $R_{dis}$ represents the longitudinal distance that the vehicle moves between two consecutive time steps, encouraging the ego vehicle to move forward. $R_{speed} = v_t / v_{max}$, where $v_t$ and $v_{max}$ denote the current velocity and the maximum velocity ($80km/h$), respectively, encouraging the ego vehicle to drive fast. $R_{yaw}$ represents the heading deviation of the ego vehicle between two consecutive time steps. $R_{steering}$ is the steering angle, guiding the vehicle to follow the correct route and reducing unnecessary steering.  $c_1$, $c_2$, $c_3$, $c_4$ are the coefficients of dense rewards $R_{dis}$, $R_{speed}$, $R_{yaw}$, $R_{steering}$, and they are set to 1.0, 0.1, -0.4, and -0.4, respectively, following the settings in \cite{yang2023model}. ${R}_{term}$ is the sparse reward which is non-zero only when the task terminates. The ego vehicle receives a +10 reward when reaching the destination and a -5 reward when departing from the lane. Since $R_{term}$ is sparse, it is not weighted by a coefficient.

\begin{figure}[tb!]
\centering
    \subfloat{\includegraphics[width=0.24\textwidth]{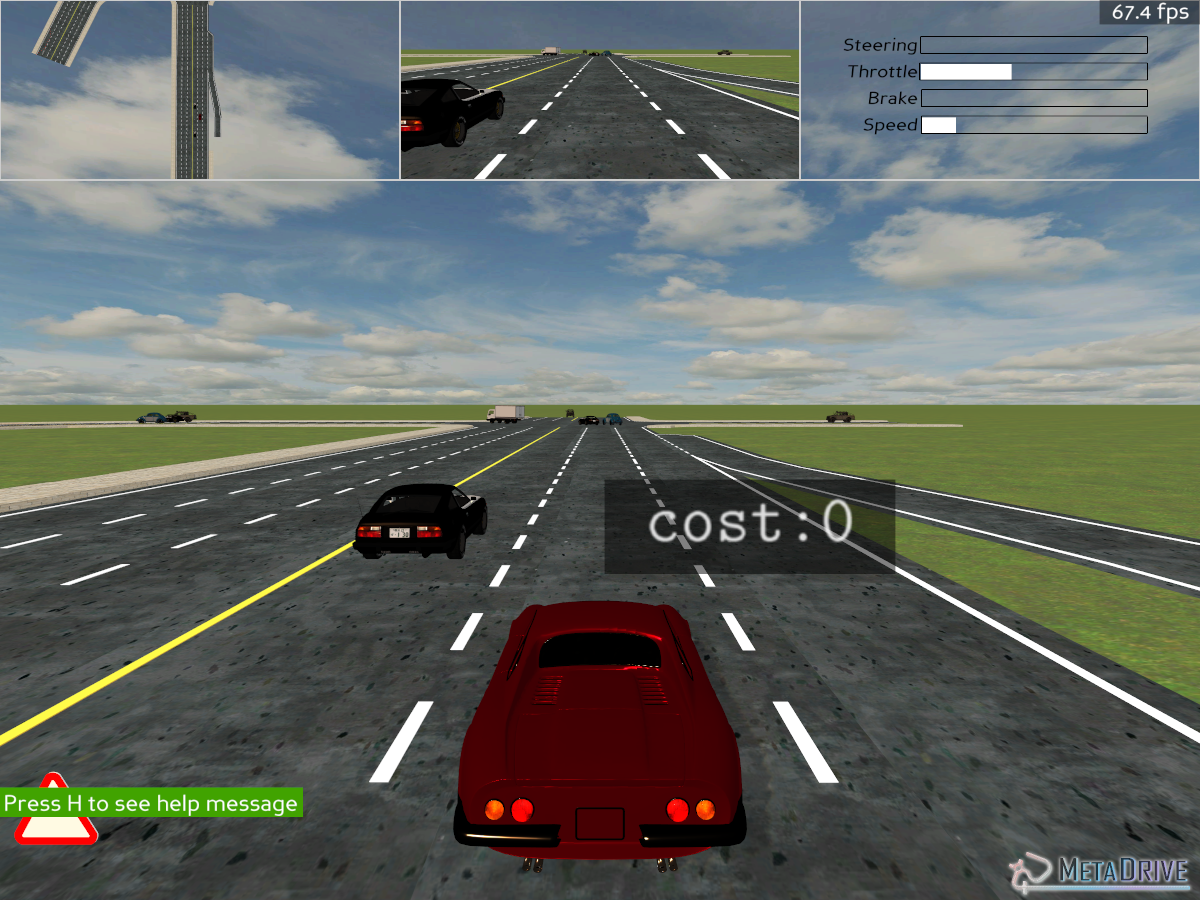}%
    \label{a}}
    \hfill
    \subfloat{\includegraphics[width=0.24\textwidth]{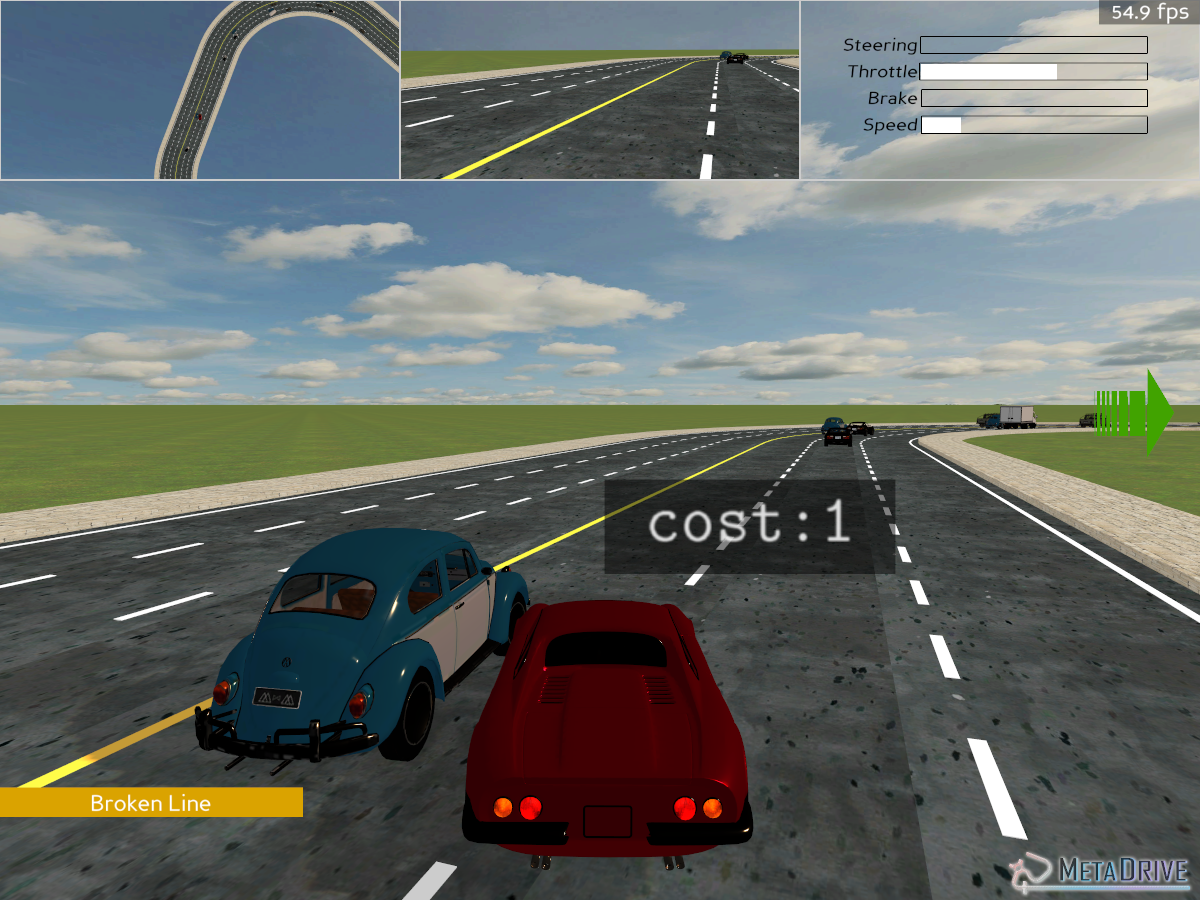}%
    \label{b}}
    \caption{MetaDrive Environment. (a) The vehicle is smoothly and safely navigating on the road with the real-time cost value of \textbf{0}. (b) A collision with another vehicle results in a cost value of \textbf{+1}.}
    \label{metacost}
\end{figure}

The cost function is the sum of some sparse cost signals, where the cost value is  \textbf{+1} when the ego vehicle encounters a collision with other objects such as vehicles, cones, roadblocks, and warning signs, etc. or departing from the lane, as shown in Fig. \ref{metacost}.

We conduct the training process using 20 road maps which are randomly generated from the elementary components. On these road maps, other randomly generated traffic vehicles have various types and common driving behaviors such as lane-changing and following. Specifically, traffic accidents will also occur at random locations on these road maps. Our evaluation is conducted on additional randomly generated 20 road maps that are unseen for the methods to be evaluated. The detailed configurations are outlined in the TABLE \ref{train}, which follows the configuration in SafeRL-Kit \cite{zhang2022saferl}. 

Our hardware platform is NVIDIA RTX 4090 paired with Intel Core i9-12900K, while the software platform is Ubuntu 20.04, with PyTorch as the deep learning framework. The hyperparameters of our method are presented in Table \ref{Hyperparameters}.

\begin{table}[tb!]
\caption{Configuration for Training and Evaluation}
\centering
\begin{tabular}{@{}lc@{}}
\toprule
\multicolumn{1}{c}{\textbf{Configuration}}           & \multicolumn{1}{l}{\textbf{Value}} \\ \midrule
The number of maps generated randomly by MetaDrive.  & 20                                 \\
The number of lanes on the road.                     & 3                                  \\
The number of vehicles per km per lane (p/km/ln).     & 12                                 \\
The average speed of other vehicles (km/h)            & 30                                 \\
The probability of accidents happening on each block & 0.8                                \\ 
\bottomrule
\end{tabular}
\label{train}
\end{table}

\begin{table}[tb!]
    \centering
    \caption{Hyperparameters}
    \begin{tabular}{cc}
    \toprule
    \textbf{KEY}& \textbf{VALUE}\\
    \midrule
    Number of hidden layers of policy-net and value-net& 2\\
    
    Hidden layer size of policy-net and value-net& 64\\

    Hidden Layers Activation of policy-net and value-net& Tanh \\

    Discount factor $\gamma$& 0.99\\

    Learning rate of policy-net and value-net& 3e-4\\
   
    Learning rate of validation-net& 3e-4\\

    Batch size& 20000\\
  
    Initial Short-term Lagrange multiplier& 0.5\\

    Initial Long-term Lagrange multiplier& 0.1\\

    Learning rate for Short-term Lagrange Multiplier& 0.01\\

    Learning rate for Short-term Lagrange Multiplier& 0.025\\

    Trajectory length $n$& 5\\
    \bottomrule
    \end{tabular}
    \label{Hyperparameters}
\end{table}


\subsection{Metrics and baselines}
Since safety is the most consideration for autonomous driving tasks, the metrics that we use to measure the vehicle safety follow those in the MetaDrive \cite{li2022metadrive}, including the success rate and episode cost. The success rate represents the ratio of the number of episodes arriving at the destination to the total number of evaluation episodes, while the episode cost is the ratio of the total cost in all episodes to the total number of evaluation episodes. Mainstream safe RL methods for AD such as \cite{zhang2022saferl}
use the two metrics to evaluate their performance. Moreover, we also use the episode reward, which is the ratio of the total reward in all episodes to the total number of evaluation episodes, as an additional metric because it can demonstrate the learning performance of RL learners. 

PPO, as a unconstrained RL learner, \cite{schulman2017proximal} is one of the comparative methods since our method is developed based on it. In order to show the advancement of our LSTC method compared with mainstream safe RL methods, the classic CMDP-based methods including PPO-Lag \cite{ray2019benchmarking} and TRPO-Lag \cite{ray2019benchmarking}, which show excellent performance in the Safty Gym \cite{ray2019benchmarking}, are also used as the comparative methods. In addition, to evaluate the performance in autonomous driving tasks, we also compare the proposed LSTC method with the SOTA methods in SafeRL-kit \cite{zhang2022saferl}, a toolkit for benchmarking efficient and safe RL methods for autonomous driving tasks. These methods include Safty Layer(QPSL) \cite{dalal2018safe}, Recover RL(REC) \cite{thananjeyan2021recovery}, Off-policy Lagrange(TD3-Lag) \cite{ha2020learning}, Feasible Actor-Critic(FAC) \cite{ma2021feasible}, and Exact Penalty Optimization(EPO) \cite{zhang2022saferl}.

\subsection{Safety performance analysis compared with SOTA methods}
To demonstrate stable results, we refer to 20 testing rounds as an experimental group. A road is randomly selected from the 20 road maps in each testing round, and each experimental group is repeated 10 times to calculate the mean and standard deviation of the success rate and episode cost as our experimental results, which are shown in TABLE \ref{mainex}.


The LSTC-based method exhibits the highest success rate, surpassing the baseline PPO by $0.2$, and it also surpasses the REC that shows the highest success rate in other comparative methods by $0.13$. The main reason lies in the duel-constraint optimization in the LSTC-based method, ensuring the vehicle to safely navigate from the start point to the destination. The PPO-Lag and TRPO-Lag methods show lower success rate compared to PPO because their constraints are only goal-oriented  that is similar to the reward-driven mechanism in RL methods, in which the vehicle is difficult to balance the two objective, leading to the failure of reaching the destination due to the reduction of exploratory performance.


\begin{table}[tb!]
\centering
\caption{Comparative results with SOTA methods}
\resizebox{0.45\textwidth}{!}{
\begin{tabular}{ccc}
\hline
\textbf{Method\textbackslash{}Metris} & success rate $\uparrow$    & episode cost $\downarrow$       \\ \hline
PPO \cite{achiam2017constrained}                                   & 0.71±0.19          & 15.25±6.11         \\ \hline
PPO-Lag \cite{ray2019benchmarking}                              & 0.63±0.12          & 2.56±0.89          \\
TRPO-Lag \cite{ray2019benchmarking}                             & 0.27±0.19          & 1.57±1.06          \\ \hline
TD3-Lag \cite{ha2020learning}                              & 0.74±0.05          & 9.23±4.88          \\
REC  \cite{thananjeyan2021recovery}                                 & 0.78±0.06          & 14.18±1.92         \\
FAC \cite{ma2021feasible}                                  & 0.68±0.04          & 3.29±0.50          \\
QPSL \cite{dalal2018safe}                                 & 0.73±0.04          & 12.91±1.10         \\ 
EPO \cite{zhang2022saferl}                                  & 0.73±0.05          & 4.29±0.71          \\ \hline
LSTC (ours)                            & \textbf{0.91±0.11} & \textbf{1.31±1.02} \\ \hline
\end{tabular}
}\label{mainex}
\end{table}

TABLE \ref{mainex} shows that our method has the lowest episode cost in all comparative methods, which means the fewest collisions and the safest decision in our method since the cost value is calculated by the numbers of collisions with other objects and departing from the lane. Compared with TRPO-Lag and FAC, the best methods in classic CMDP and SafeRL-kit, our LSTC method achieves lower episode cost values by $0.26$ and $1.98$, respectively. The main reason is that the short-term constraint in the LSTC method considers the vehicle's state-based safety, which encompasses avoidance of collisions, so as to decrease the episode cost and improve the safety performance. In contrast, other safe RL methods rely on the goal-oriented constraint that only considers the long-term safety, achieving higher  episode cost values. Since PPO doesn't consider risks and aims to solely maximize the expected reward, its episode cost is the highest in all the comparative methods. The episode cost of REC is also much higher than our method because it learns the risk policy to minimize the cost. However, the risk policy takes over the main policy in hazardous states, which it struggles to learn.

\subsection{Comparative results in different complex scenarios}
In AD tasks, some scenarios pose significant challenges such as inputting ramp (InRamp), sharp left-turn (Slet), roundabout (Rodab), and intersection (Intes), so we conduct additional experiments to evalute our method in these scenarios. 

\begin{figure*}
  \centering
  \begin{tabular}{ccccc}
     & PPO & PPO-Lag & EPO & LSTC (ours) \\
     \rotatebox[origin=c]{90}{inputting ramp}                         & \adjustbox{valign=m}{\includegraphics[width=0.20\textwidth]{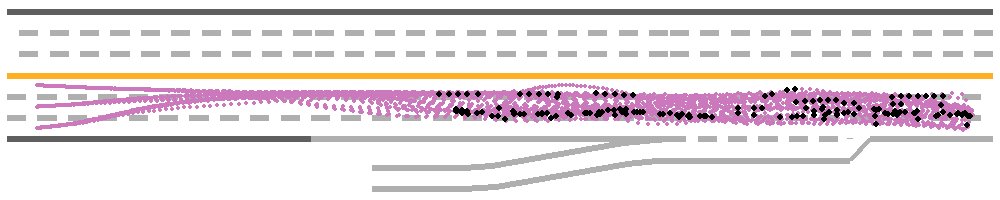}}   & \adjustbox{valign=m}{\includegraphics[width=0.20\textwidth]{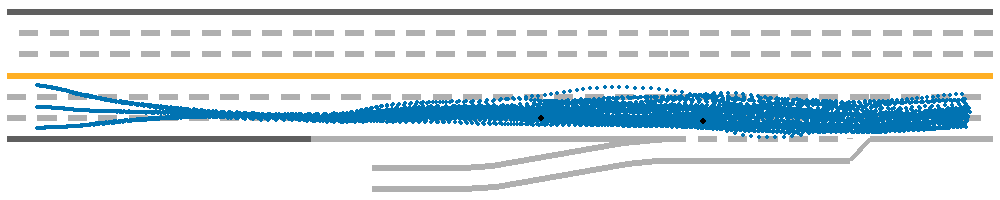}}      & \adjustbox{valign=m}{\includegraphics[width=0.20\textwidth]{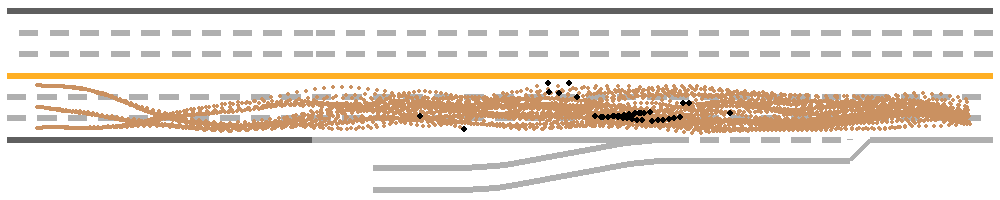}}   & \adjustbox{valign=m}{\includegraphics[width=0.20\textwidth]{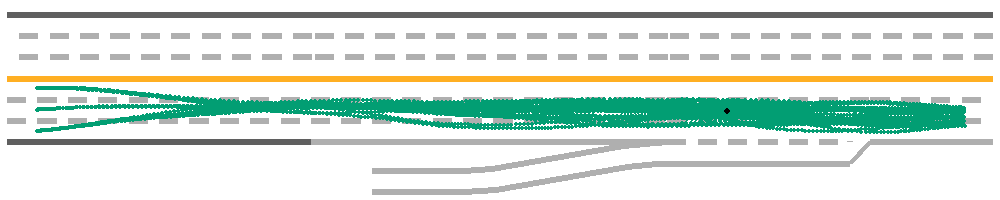}}    \\[1.0cm]
    \addlinespace
    \rotatebox[origin=c]{90}{sharp left-turn}                  & \adjustbox{valign=m}{\includegraphics[width=0.20\textwidth]{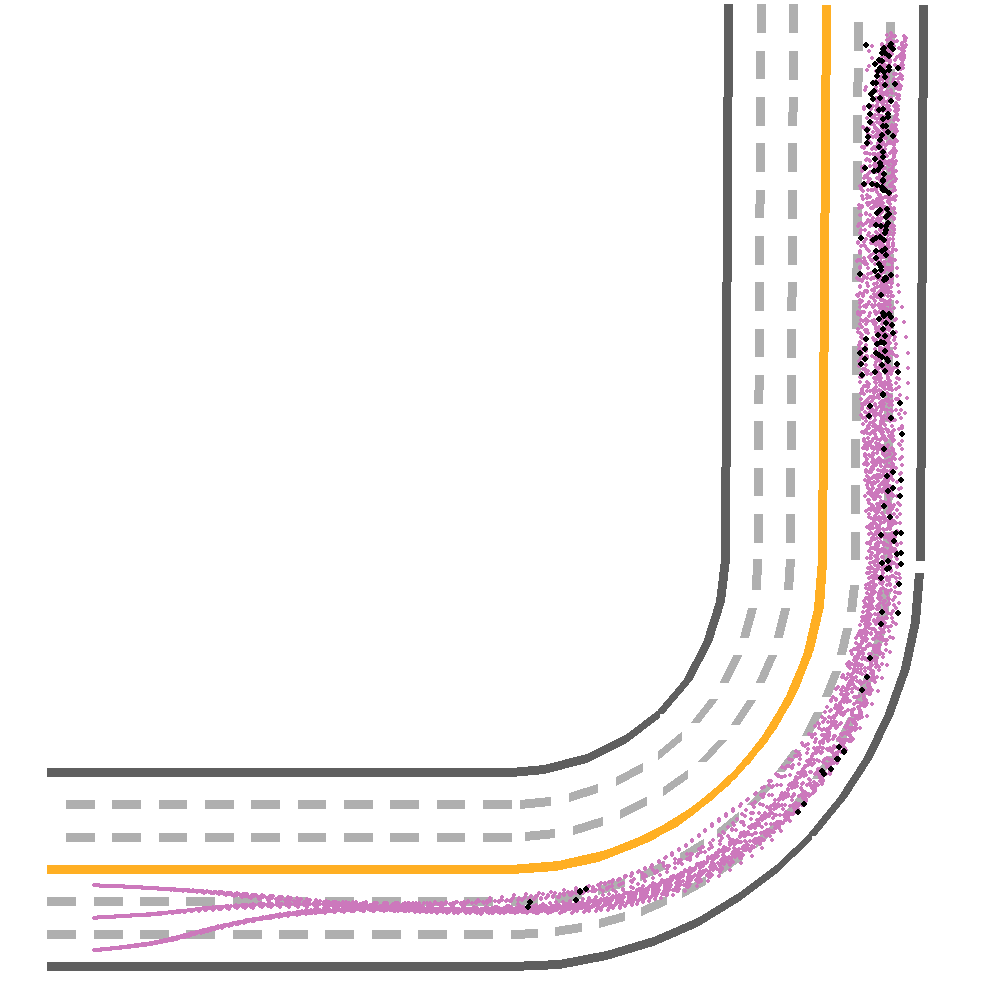}}   & \adjustbox{valign=m}{\includegraphics[width=0.20\textwidth]{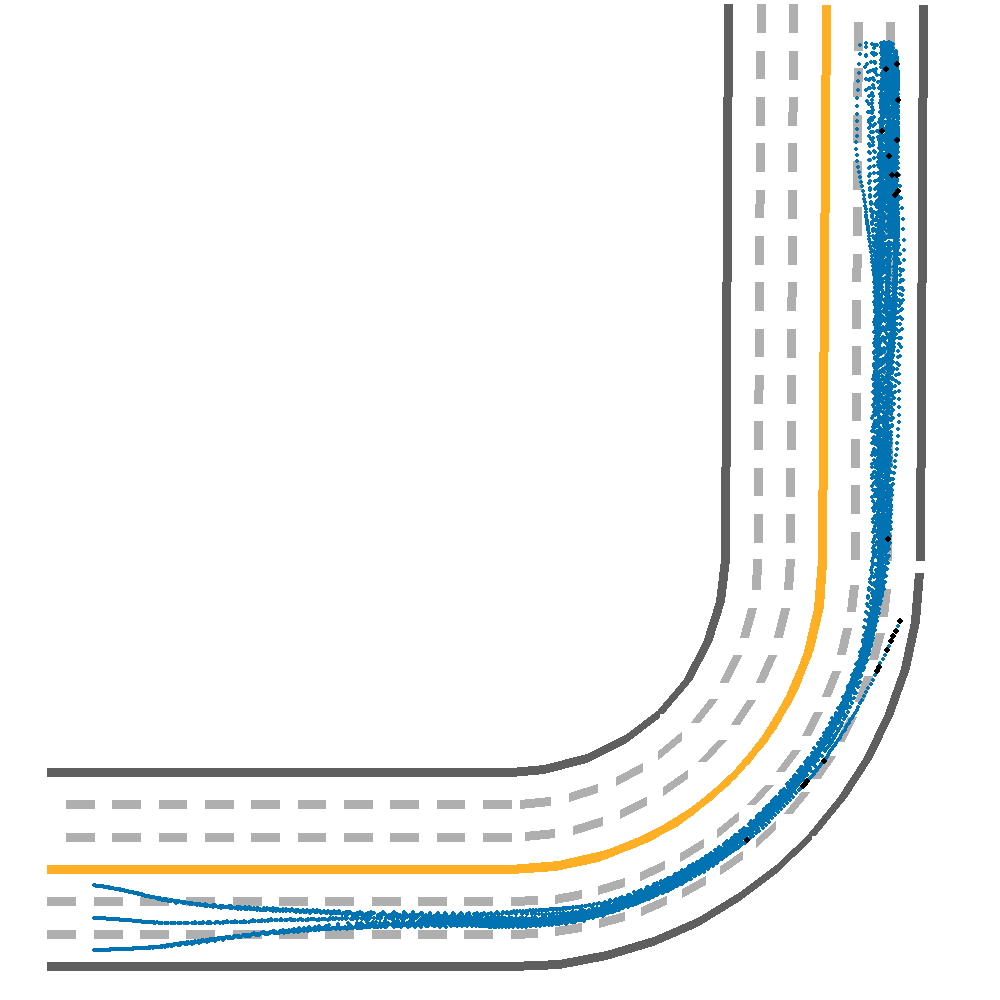}}      & \adjustbox{valign=m}{\includegraphics[width=0.20\textwidth]{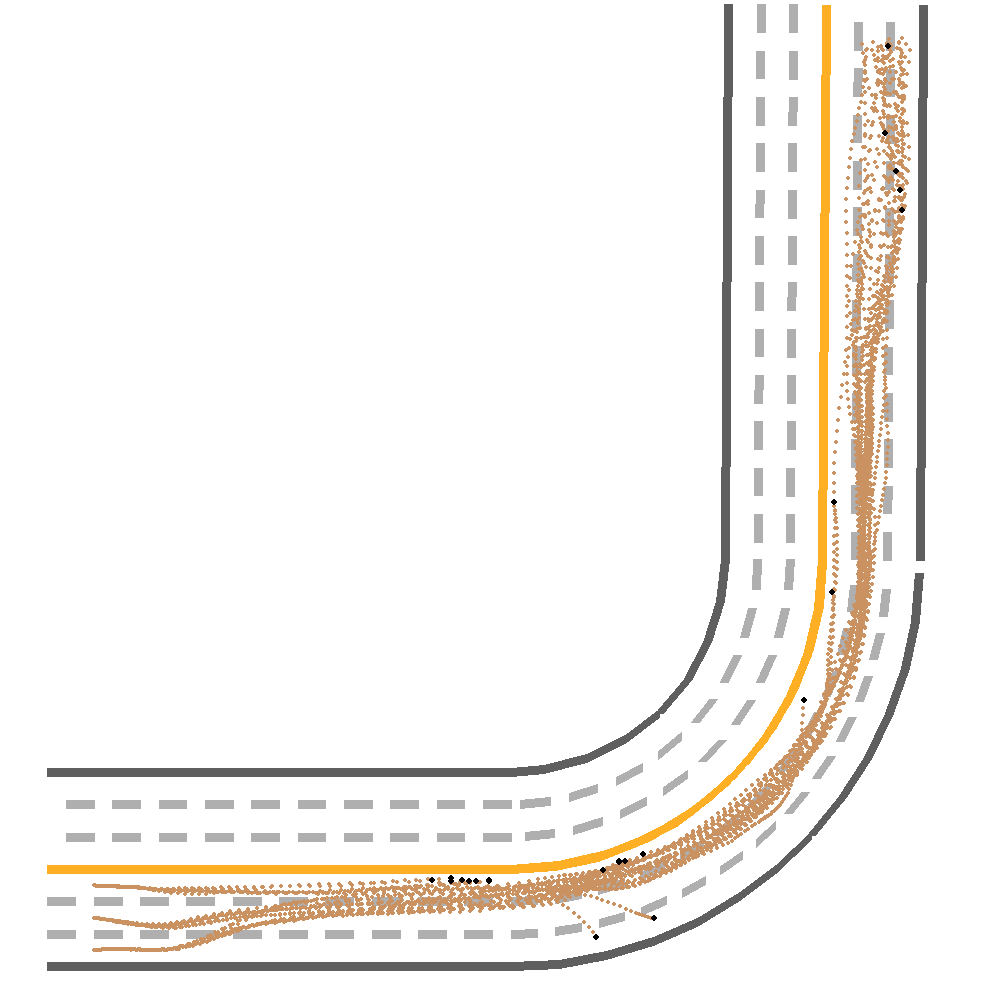}}   & \adjustbox{valign=m}{\includegraphics[width=0.20\textwidth]{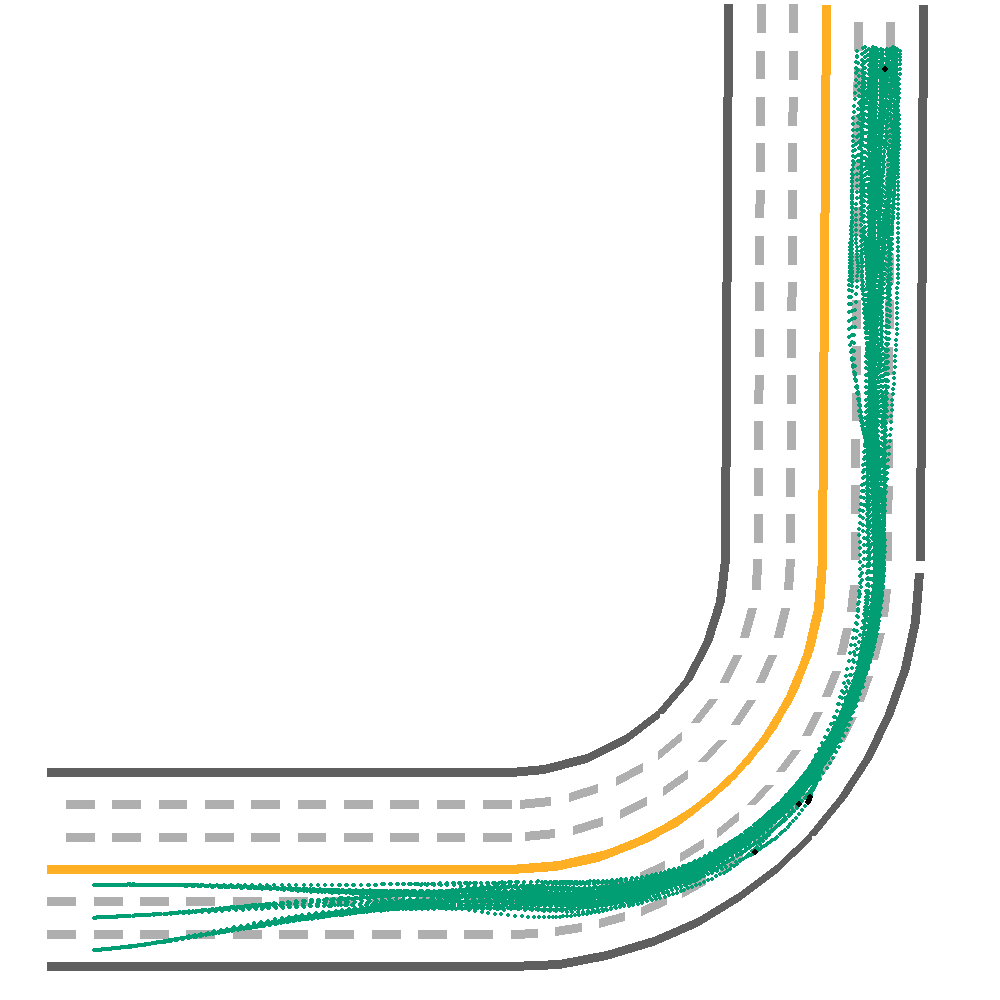}}    \\[1.0cm]
    \addlinespace
    \rotatebox[origin=c]{90}{roundabout}                       & \adjustbox{valign=m}{\includegraphics[width=0.20\textwidth]{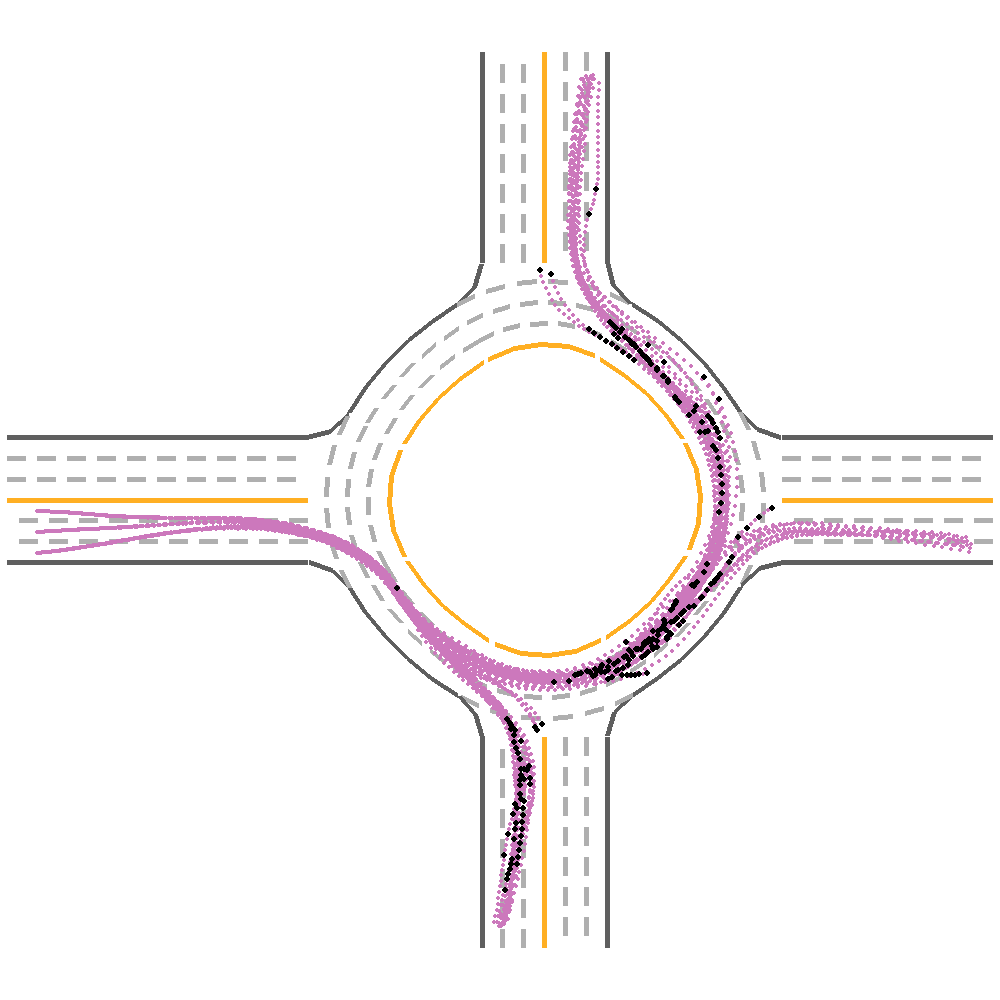}}   & \adjustbox{valign=m}{\includegraphics[width=0.20\textwidth]{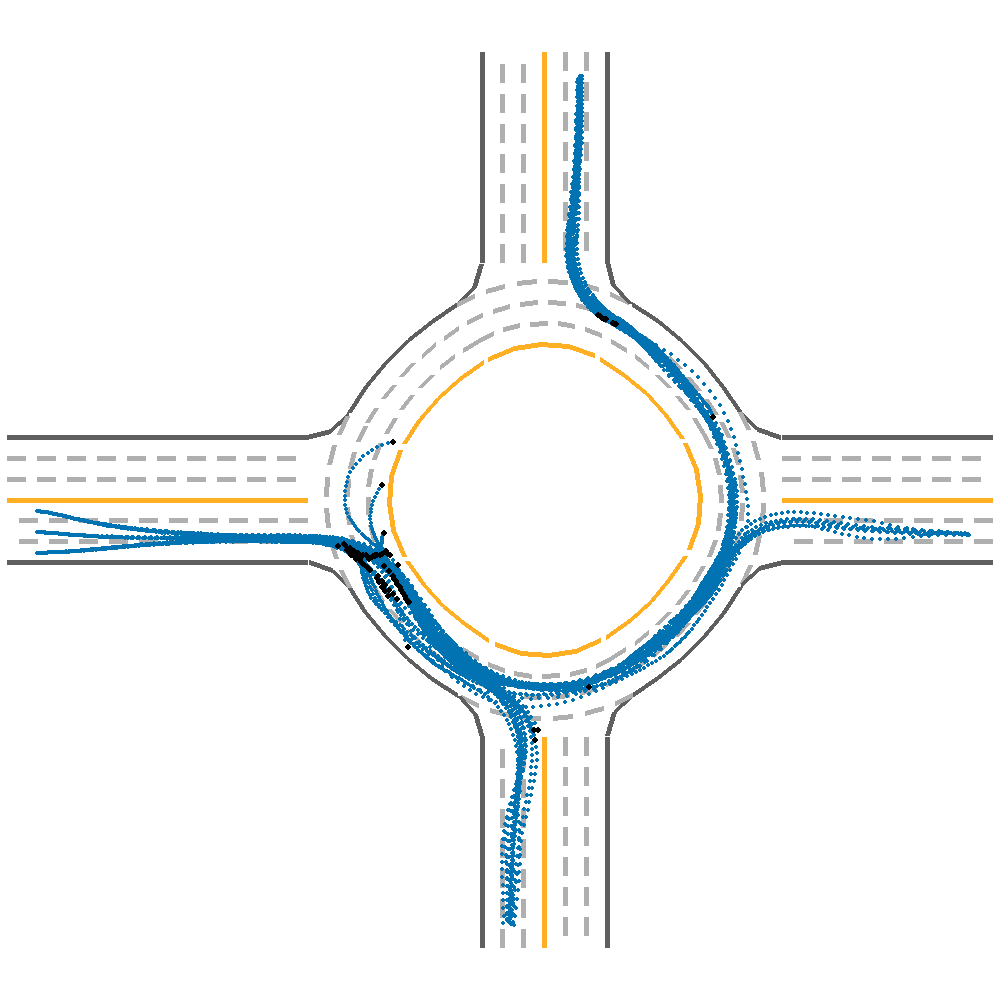}}      & \adjustbox{valign=m}{\includegraphics[width=0.20\textwidth]{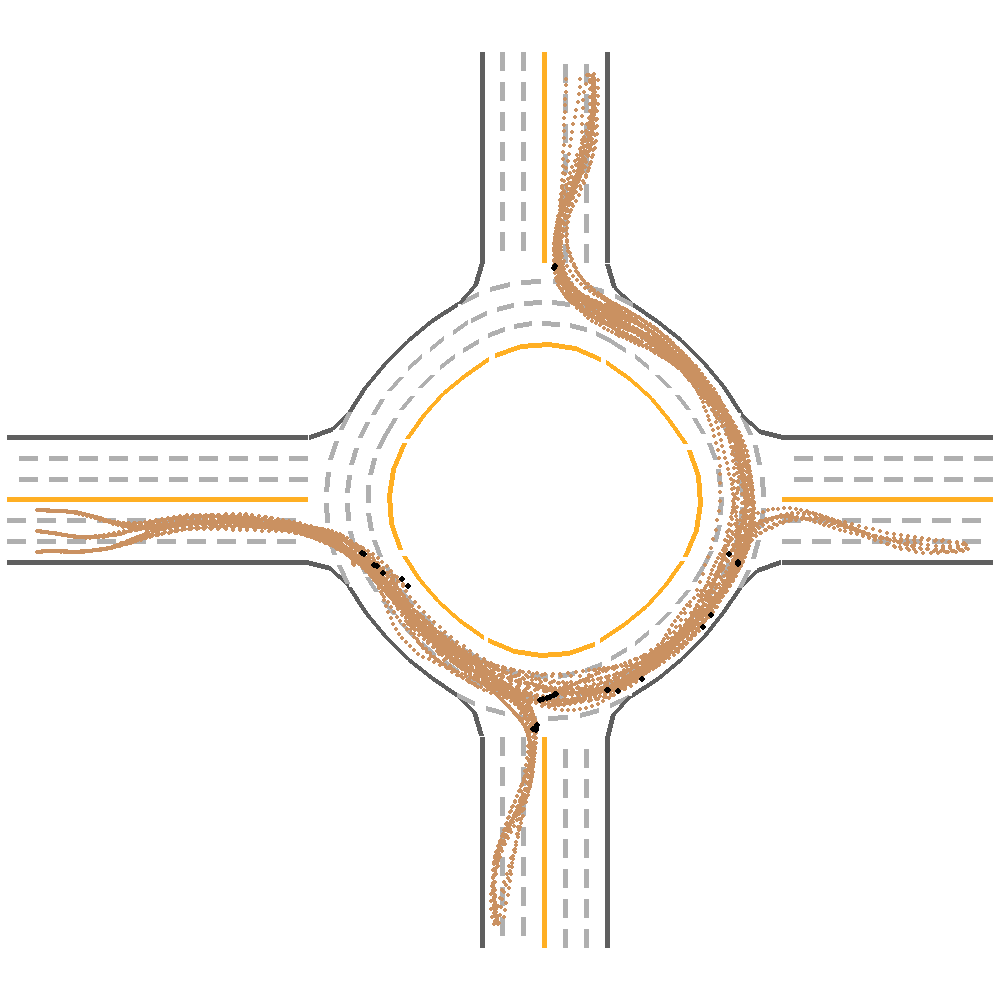}}   & \adjustbox{valign=m}{\includegraphics[width=0.20\textwidth]{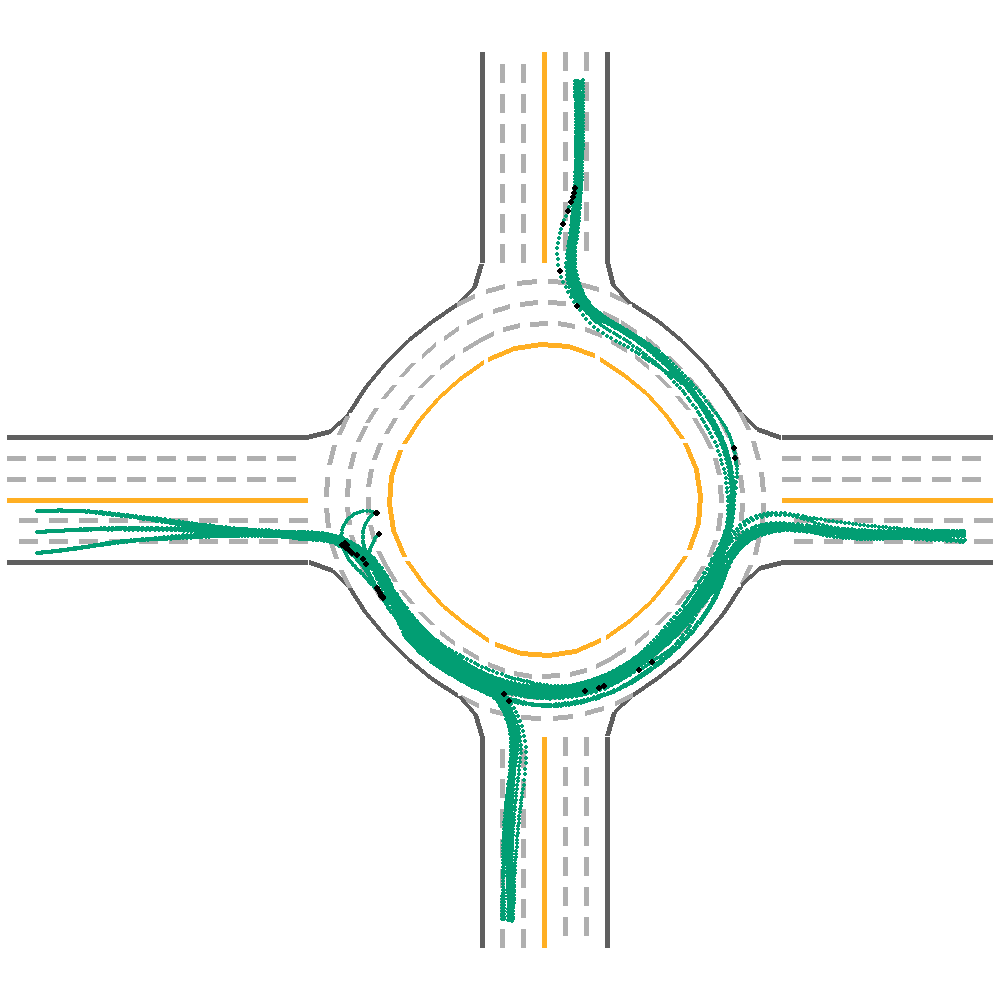}}    \\[1.0cm]
    \addlinespace
    \rotatebox[origin=c]{90}{intersection}                     & \adjustbox{valign=m}{\includegraphics[width=0.20\textwidth]{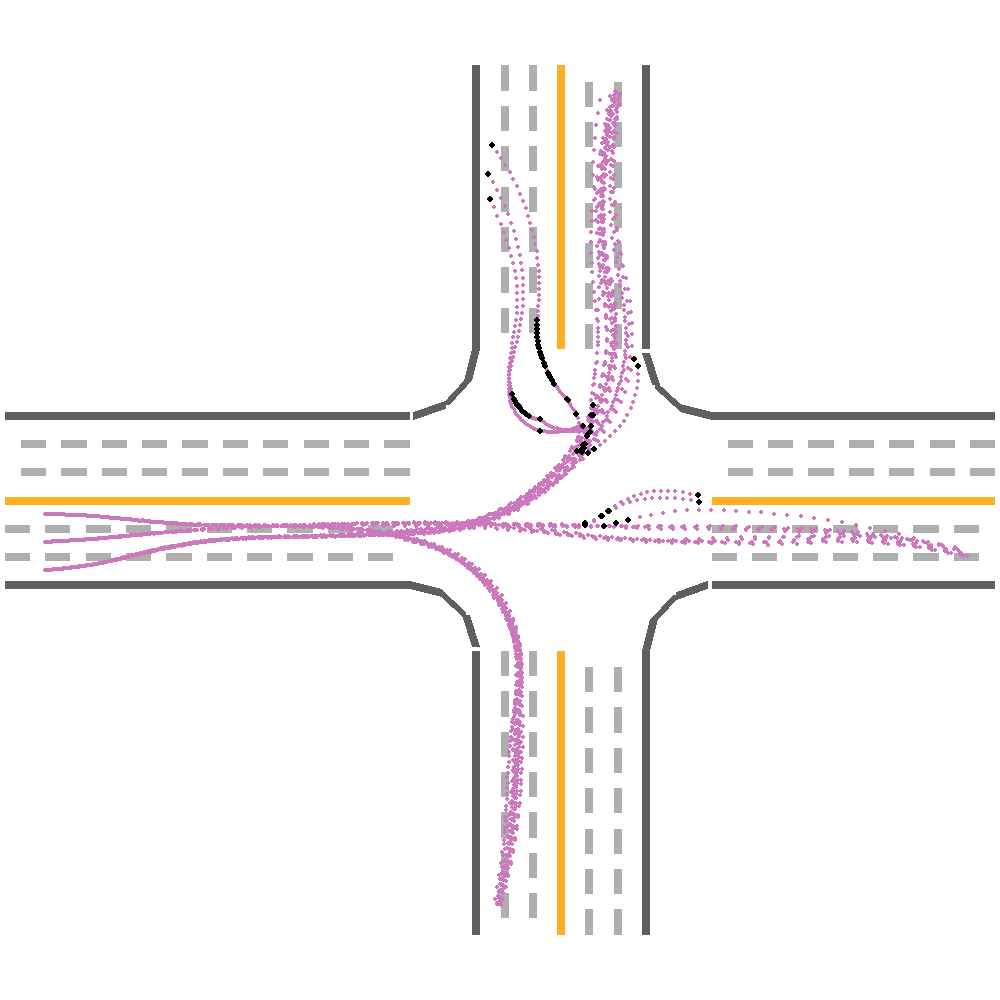}}   & \adjustbox{valign=m}{\includegraphics[width=0.20\textwidth]{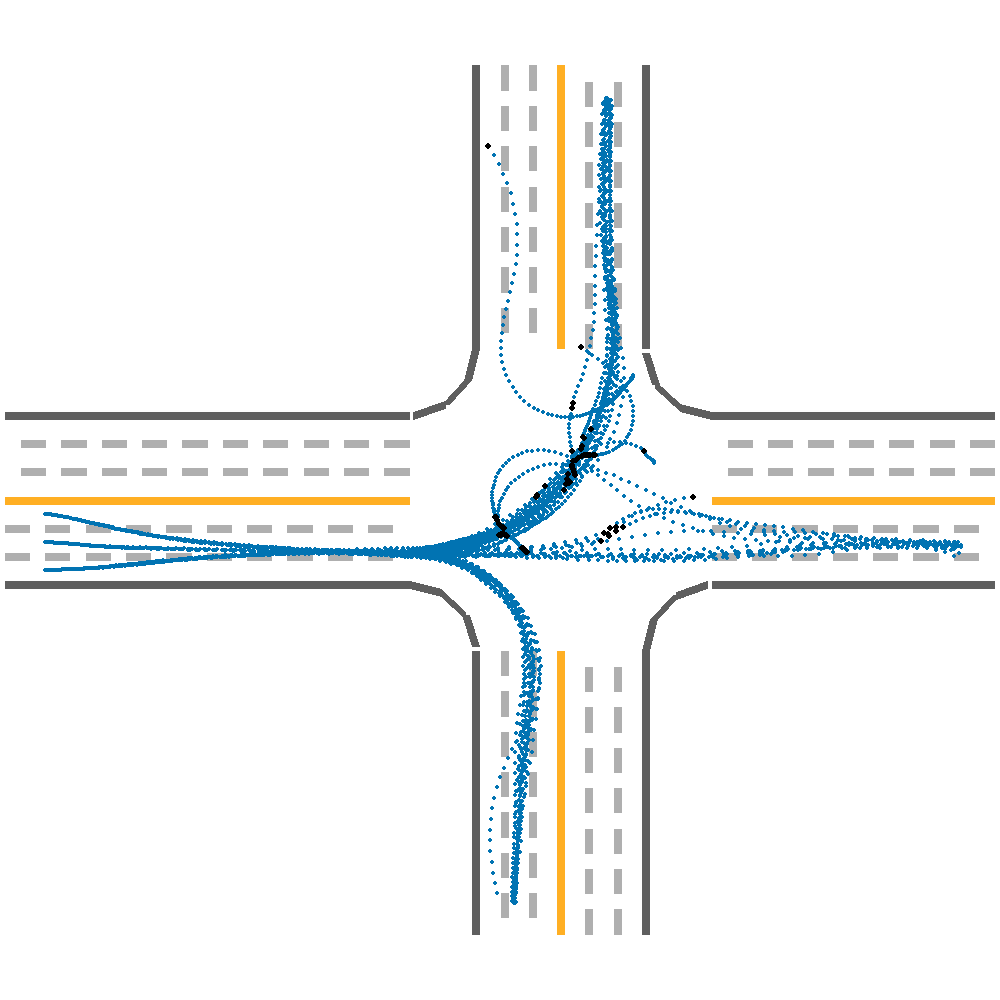}}      & \adjustbox{valign=m}{\includegraphics[width=0.20\textwidth]{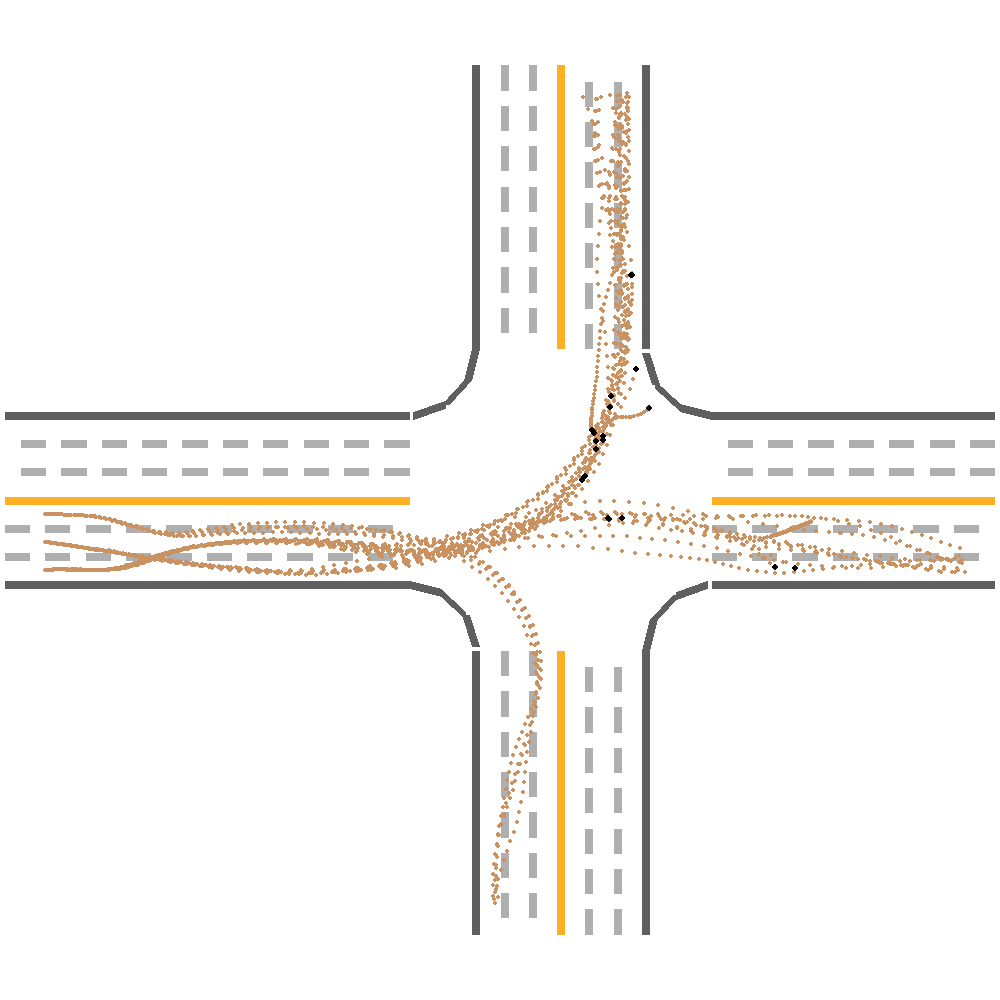}}   & \adjustbox{valign=m}{\includegraphics[width=0.20\textwidth]{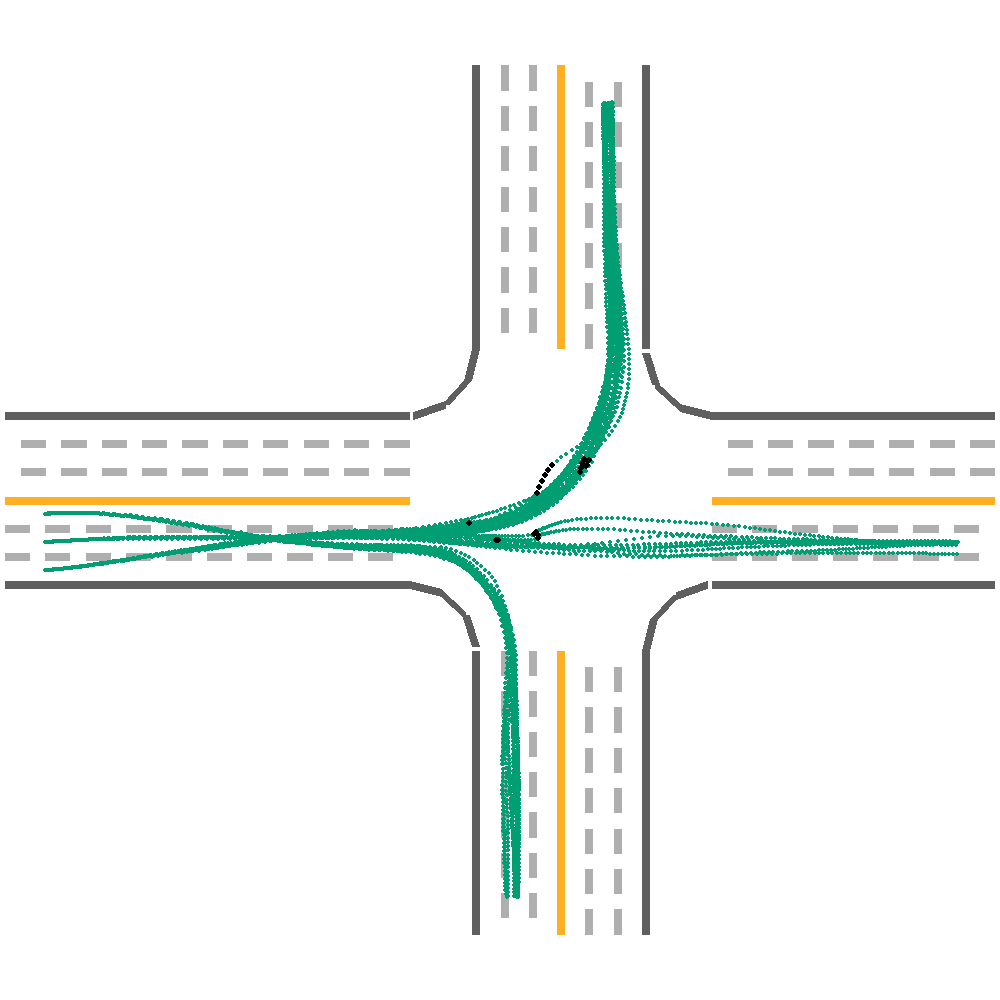}}   
  \end{tabular}
  \caption{Driving trajectories in different scenarios. Sparse points in the trajectories denote the vehicle at fast speeds, and vice versa.}
  \label{complexex}
\end{figure*}

We select the best method from each of the three kinds of methods including the unconstrained RL method, classic CMDP-based RL method, and safe RL method in SafeRL-kit, which are the PPO, PPO-Lag, and EPO, respectively, as the representative comparative methods. Each selected method is evaluated 50 times in the four scenarios. The ego vehicles randomly choose start and exit points on the roads with three lanes. Other traffic vehicles and static obstacles such as cones, roadblocks, and warning signs are randomly placed on the road. The success rates and average costs are shown in TABLE \ref{complextb}. Additionally, we record the vehicle's trajectories and the coordinates where accidents occur in all the episodes, which are as shown in Fig. \ref{complexex}. 

\begin{table*}[h]
\centering
\caption{Comparative results in different scenarios}
\resizebox{0.68\textwidth}{!}{\begin{tabular}{@{}c|cccc|cccc@{}}
\toprule
metrics   & \multicolumn{4}{c|}{success rate$\uparrow$}                                 & \multicolumn{4}{c}{episode cost$\downarrow$}                                    \\ 
scenarios & InRamp            & Slet            & Rodab           & Intes            & InRamp           & Slet           & Rodab           & Intes           \\  \midrule
PPO \cite{achiam2017constrained}      & 98\%           & \textbf{100\%} & 80\%          & 86\%           & 3.10          & 3.70          & 4.70         & 1.32          \\
PPO-Lag \cite{ray2019benchmarking}    & 98\%           & 78\%           & 78\%          & 92\%           & 0.04          & 1.82          & 1.82          & 1.82          \\
EPO  \cite{zhang2022saferl}     & 94\%           & 52\%           & 72\%          & 86\%           & 0.88          & 0.58          & \textbf{0.60}          & \textbf{0.46} \\
LSTC (ours)      & \textbf{100\%} & \textbf{100\%} & \textbf{94\%} & \textbf{100\%} & \textbf{0.02} & \textbf{0.12} & 0.88 & 0.64          \\ \bottomrule
\end{tabular}
}
\label{complextb}
\end{table*}

TABLE \ref{complextb} shows that our method exhibits the highest success rates in all the scenarios and the lowest episode costs in the inputting ramp and sharp left-turn scenarios. According to the trajectory points shown in Fig. \ref{complexex}, The trajectory points of LSTC are evenly distributed across all scenarios, indicating that the vehicle maintains a consistent speed and mainly navigates in the middle lane with minimal lane changes. The trajectory points of PPO are the sparsest of all, which indicates that the ego vehicle consistently maintains a high-speed navigation and disregards potential dangers. EPO uses the state-wise constraints for the cumulative cost as same as FAC to ensure strict safety guarantee, but EPO reduces the state-dependent Lagrange multipliers in FAC to a single fixed hyper-parameter. In this case, EPO achieves an overall lower episode cost, but its success rate is lower than other methods because the vehicle often chooses to depart from the lane to avoid collisions with other vehicles. In the roundabout and intersection scenarios, the episode costs of our method are a slightly higher than those of EPO. The  increased costs come from the collisions that mainly occur at the entrances and centers of roundabouts and intersections because the decisions made by LSTC are more cautious compared with EPO, leading to  collides with other non-intelligent vehicles for the ego vehicle in LSTC when waiting to yield.

\subsection{Learning performance analysis}
Learning performance is a significant aspect for RL methods. The baseline PPO method and classic safe RL methods including PPO-Lag and TRPO-Lag are selected as the comparative methods to evaluate the learning performance. Besides the success rate and episode cost, we add two additional metrics, the episode reward and feasible state rate, to analyze the learning performance. The episode reward can demonstrate the learning performance of RL learners. The feasible state rate signifies the proportion of feasible states among the sampled states in an epoch, so as to quantify the exploration space of the vehicle. The total number of training time steps is $1\times 10^{7}$.
\begin{figure}[tb!]
    \centering
    \includegraphics[width=0.8\linewidth]{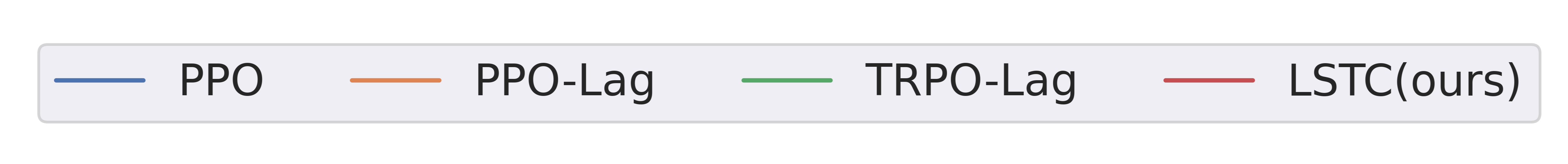}
    \subfloat{\includegraphics[width=0.48\linewidth]{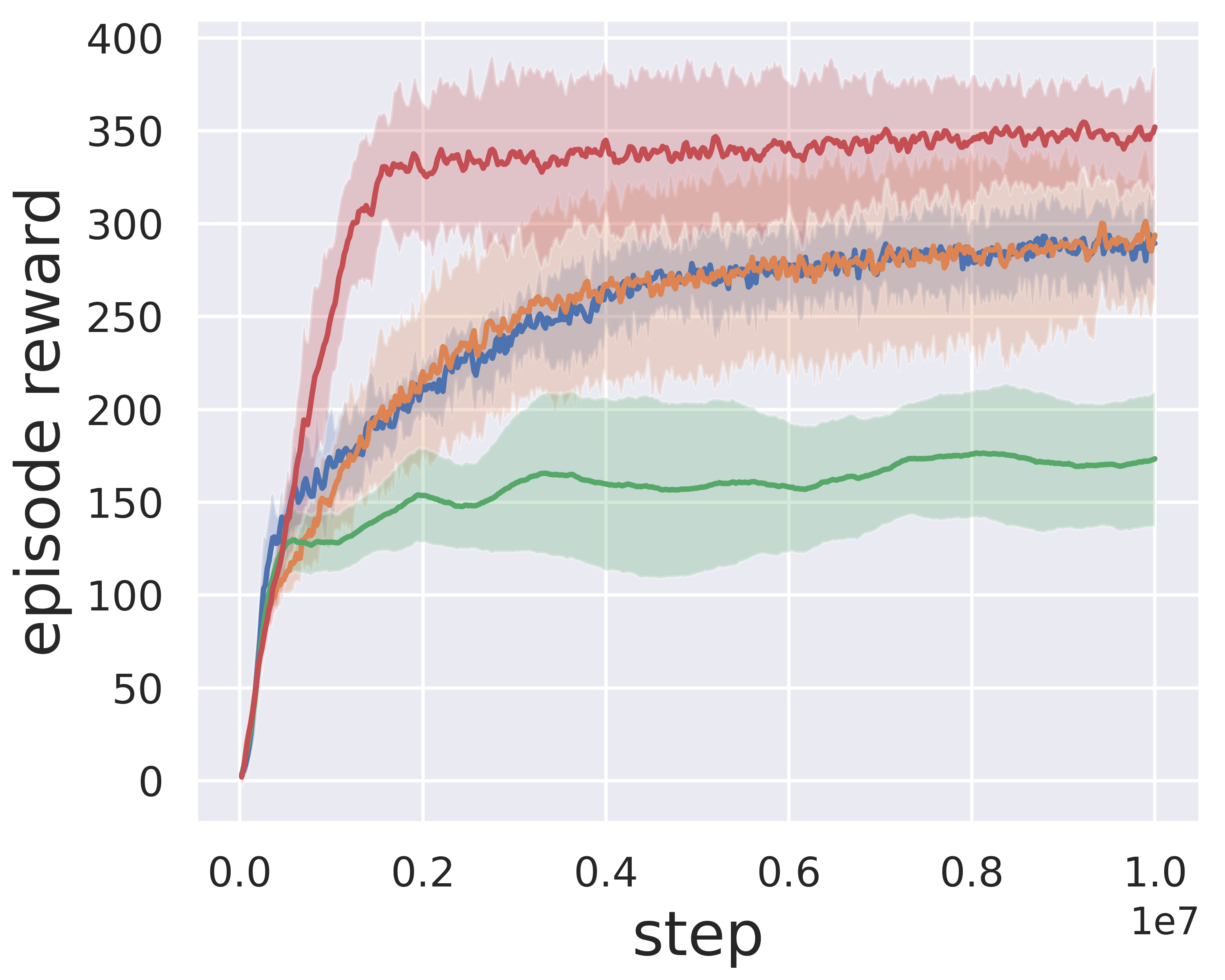}%
    \label{tain:a}}
    \subfloat{\includegraphics[width=0.48\linewidth]{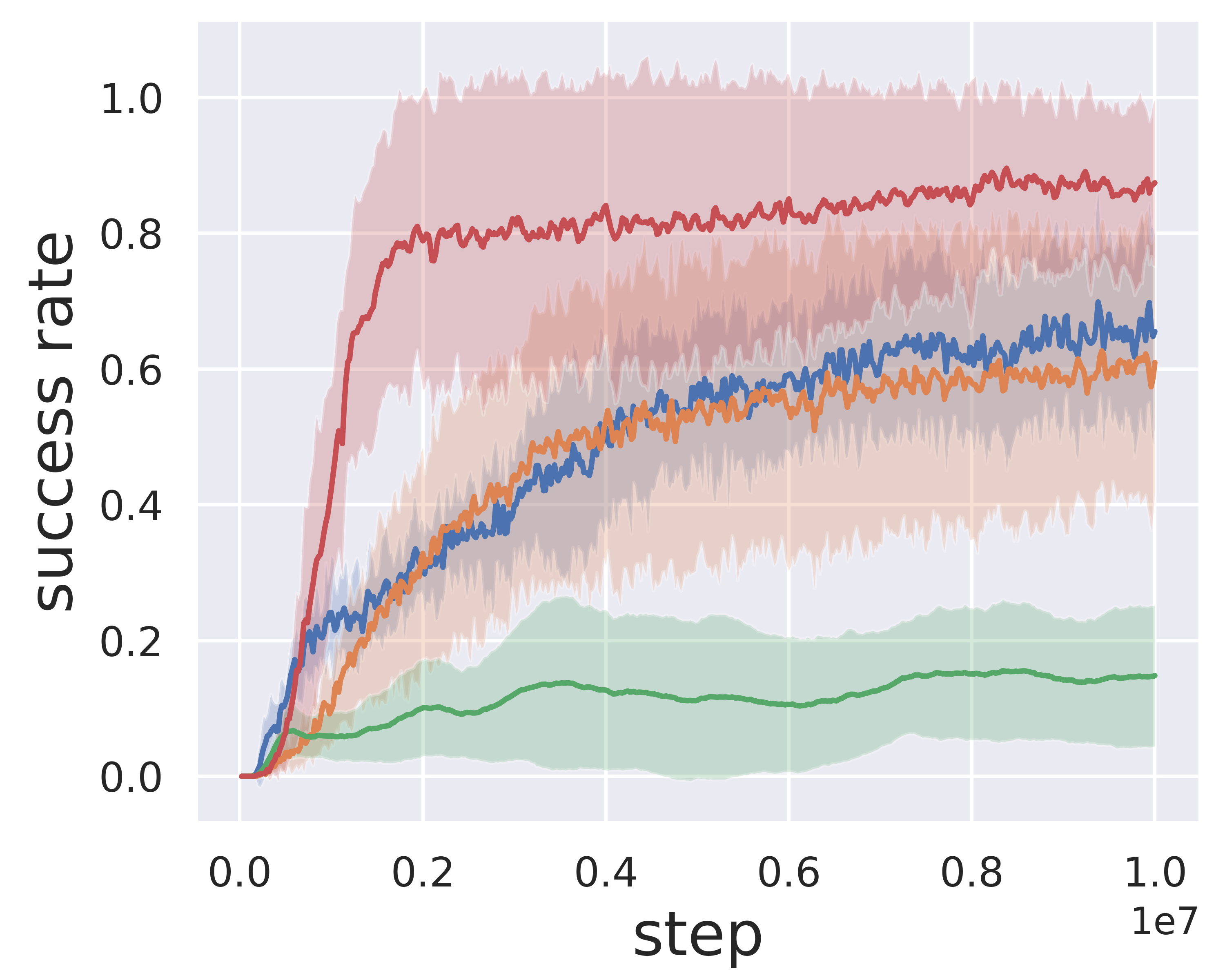}%
    \label{tain:b}} \\
    \subfloat{\includegraphics[width=0.48\linewidth]{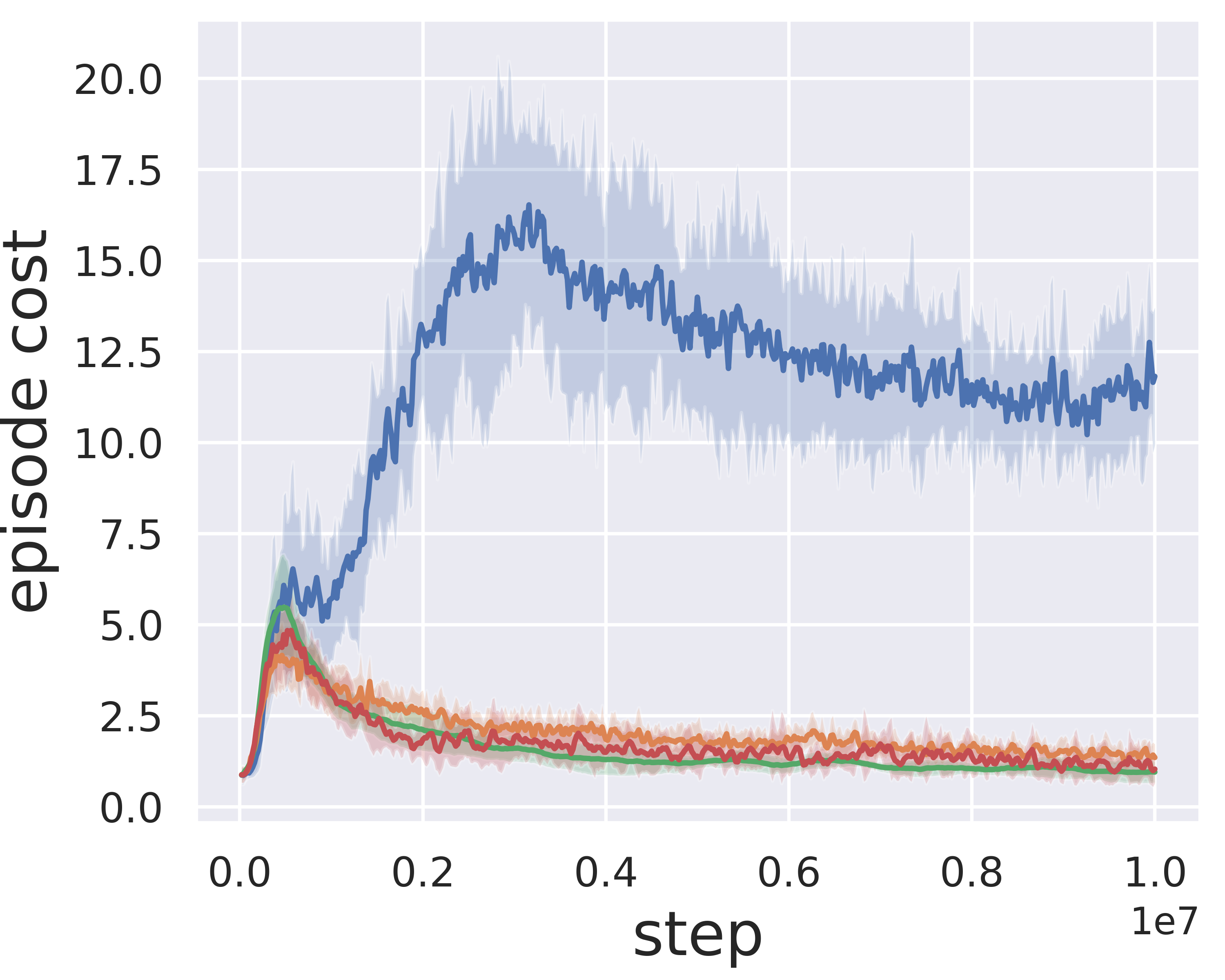}%
    \label{tain:c}}
    \subfloat{\includegraphics[width=0.48\linewidth]{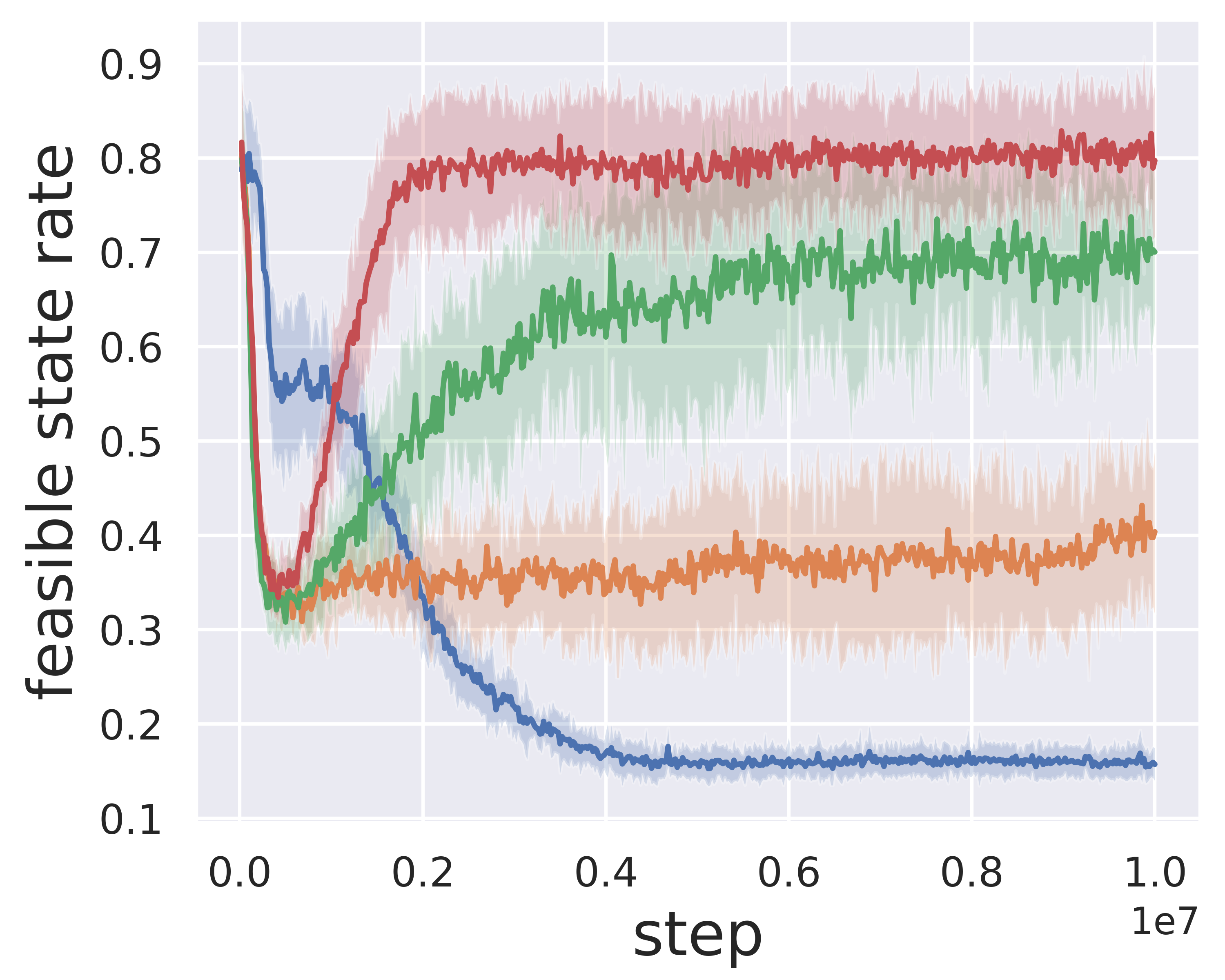}%
    \label{tain:d}}
    \caption{Training process curves. (a) Step-episode reward curves. (b) Step-success rate curves. (c) Step-episode cost curves. (d) Step-feasible state rate curves}
    \label{trainfig}
\end{figure}

 The training curves are illustrated in Fig. \ref{trainfig}. The proposed LSTC method performs best in terms of the convergence speed and training stability, as shown in Fig.\ref{trainfig}. Its episode reward converges to around 350 at $1.5\times 10^{6}$ steps and exceeds both PPO-Lag and TRPO-Lag, or even the unconstrained method PPO which has theoretically the fastest learning speed due to the simple algorithm. Moreover, safety constraints aim to reduce collisions and ensure the vehicle safety, and the short-term constraint in LSTC is used to constrain the exploration within the feasible space, so LSTC rapidly converges to a low episode cost with a high success rate and a high feasible state rate, which indicates that the LSTC method is a robust end effective learner for autonomous driving. In contrast, the episode cost of PPO remains above 10 throughout the entire training, as it doesn't consider the risk in the driving process. Since PPO-Lag and TRPO-Lag only consider the long-term constraint without the state-based constraint, vehicles often get stuck in potentially hazardous states during training, which leads to higher episode costs and lower success rates.

\section{Conclusions}
To handle the issue of unsafe state occurrence in the training process for existing safe RL methods, a long and short-term constraints-based safe reinforcement learning method is proposed for autonomous driving in this paper. We first propose the long and short-term constraints for safe RL to address the issue of unrestricted exploration, and then develop a safe RL method with dual-constraint optimization for autonomous driving to address the constrained optimization problems. By combining the long and short-term constraints, the proposed method effectively restricts the vehicle exploration within the safe space to ensure the vehicle safety. Experiments on the MetaDrive simulator show that our method outperforms both traditional CMDP-based and existing safe RL methods in continuous control tasks for autonomous driving in complex scenarios and exhibits superior safety and learning performance compared to SOTA methods. Future work will focus on addressing the issue of the dynamic sequence length of the short-term constraint for different scenarios to further improve the safety and learning performance.

\ifCLASSOPTIONcaptionsoff
  \newpage
\fi

\bibliographystyle{IEEEtran} 
\bibliography{SafeRLAD.bib} 

\begin{IEEEbiography}[{\includegraphics[width=1in,height=1.25in,clip,keepaspectratio]{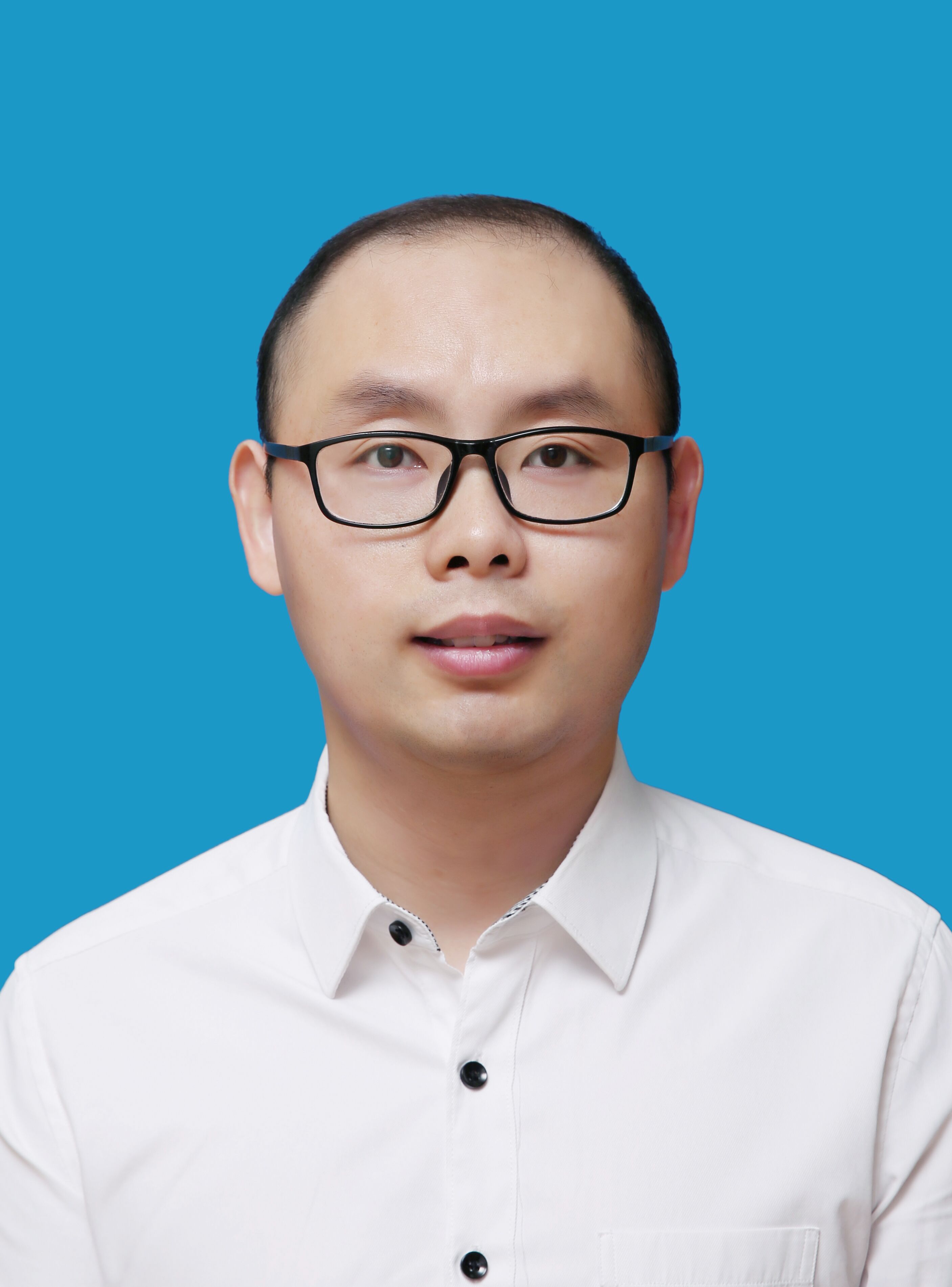}}]
	{Xuemin Hu}
	is currently a Professor with School of Artificial Intelligence, Hubei University, Wuhan, China. He received the B.S. degree in Biomedical Engineering from Huazhong University of Science and Technology and the Ph.D. degree in Signal and Information Processing from Wuhan University in 2007 and in 2012, respectively. He was a visiting scholar in the University of Rhode Island, Kingston, RI, US from November 2015 to May 2016. His areas of interest include computer vision, machine learning, motion planning, and autonomous driving. 
\end{IEEEbiography}
\vspace{-10 mm}

\begin{IEEEbiography}[{\includegraphics[width=1in,height=1.25in,clip,keepaspectratio]{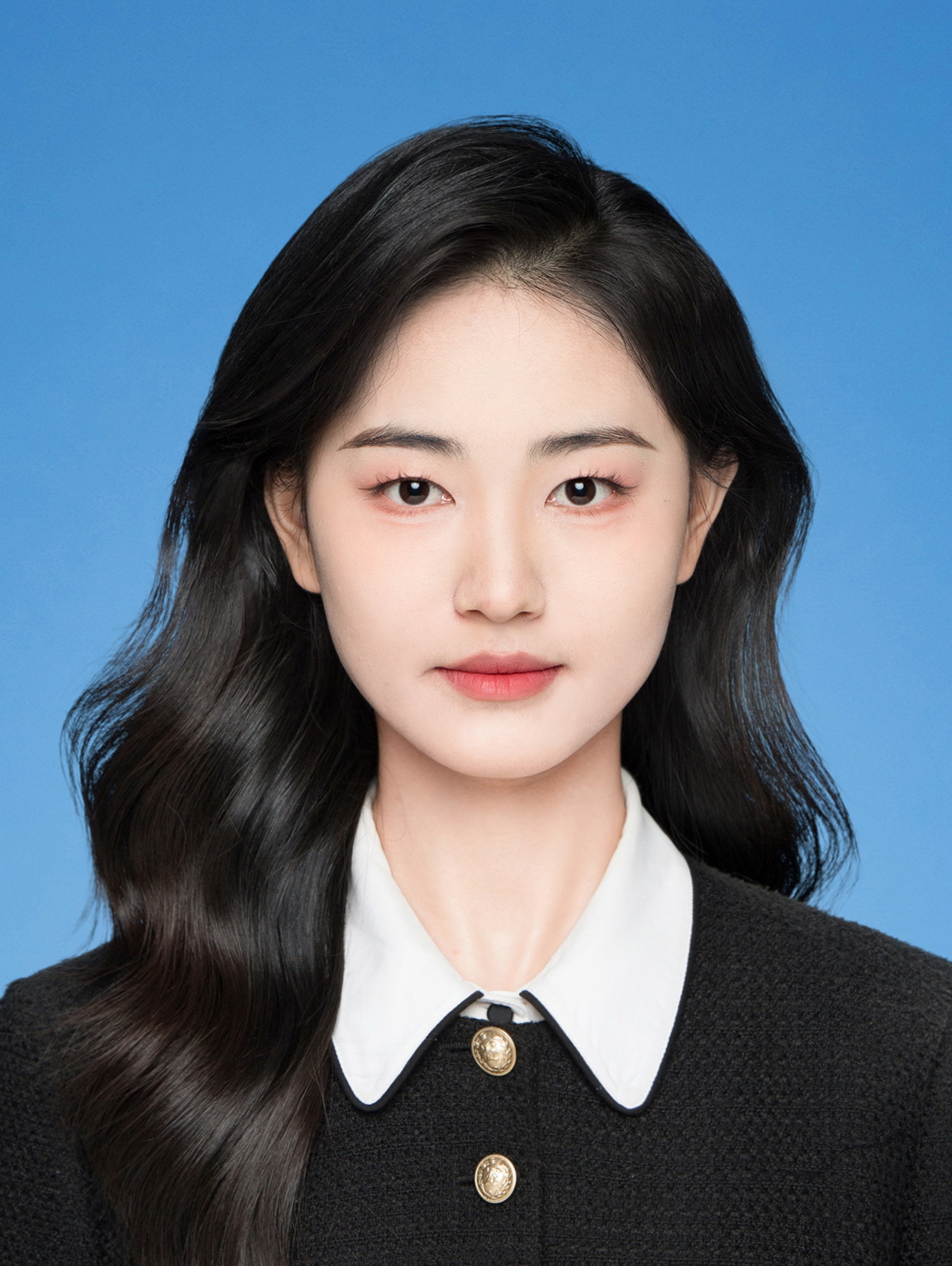}}]
	{Pan Chen}
	received the B.S. degree in Electronic Information Engineering from Hunan University of Humanities and Technology in 2023. From September 2023 to now, she is pursuing her Master's degree in School of Artificial Intelligence, Hubei University, Wuhan, China. Her areas of interest include reinforcement learning and autonomous driving.
\end{IEEEbiography}
\vspace{-10 mm}

\begin{IEEEbiography}[{\includegraphics[width=1in,height=1.25in,clip,keepaspectratio]{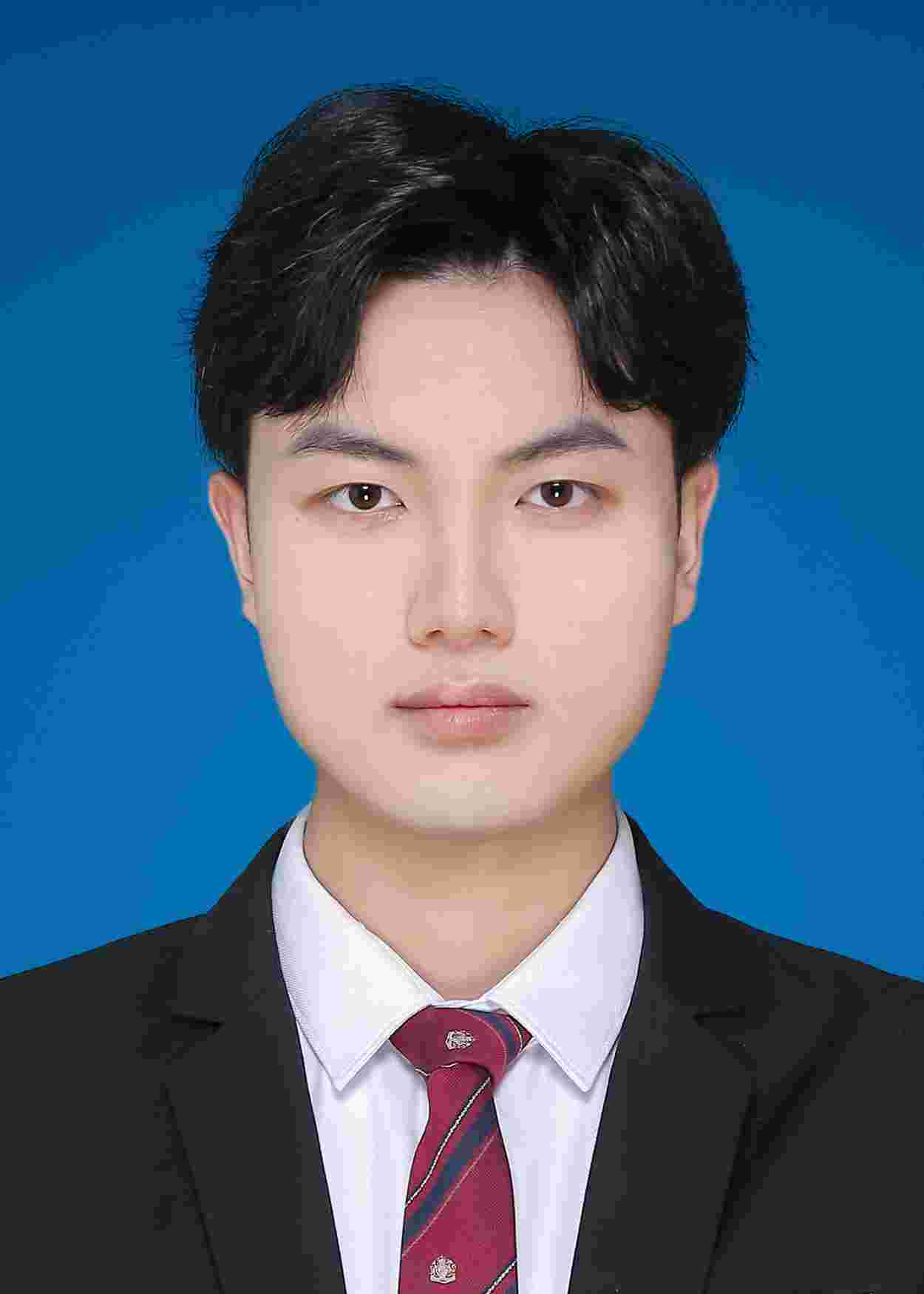}}]
	{Yijun Wen}
	received the B.S. degree in Communication Engineering from Hubei University in 2021. From September 2021 to now, he is pursuing his Master's degree in School of Artificial Intelligence, Hubei University, Wuhan, China. His areas of interest include reinforcement learning and autonomous driving.
\end{IEEEbiography}
\vspace{-10 mm}

\begin{IEEEbiography}[{\includegraphics[width=1in,height=1.25in,clip,keepaspectratio]{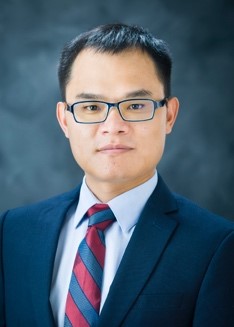}}]
	{Bo Tang}
	is an Associate Professor in the Department of Electrical and Computer Engineering at Worcester Polytechnic Institute. Prior to this, he was an Assistant Professor in the Department of Electrical and Computer Engineering at Mississippi State University. He received the Ph.D. degree in electrical engineering from University of Rhode Island (Kingstown, RI) in 2016. His research interests lie in the general areas of bio-inspired artificial intelligence (AI), AI security, edge AI, and their applications in Cyber-Physical Systems (e.g., wireless networks, autonomous vehicles, and power systems). He is currently an Associate Editor for IEEE Transactions on Neural Networks and Learning Systems.
\end{IEEEbiography}
\vspace{-10 mm}

\begin{IEEEbiography}[{\includegraphics[width=1in,height=1.25in,clip,keepaspectratio]{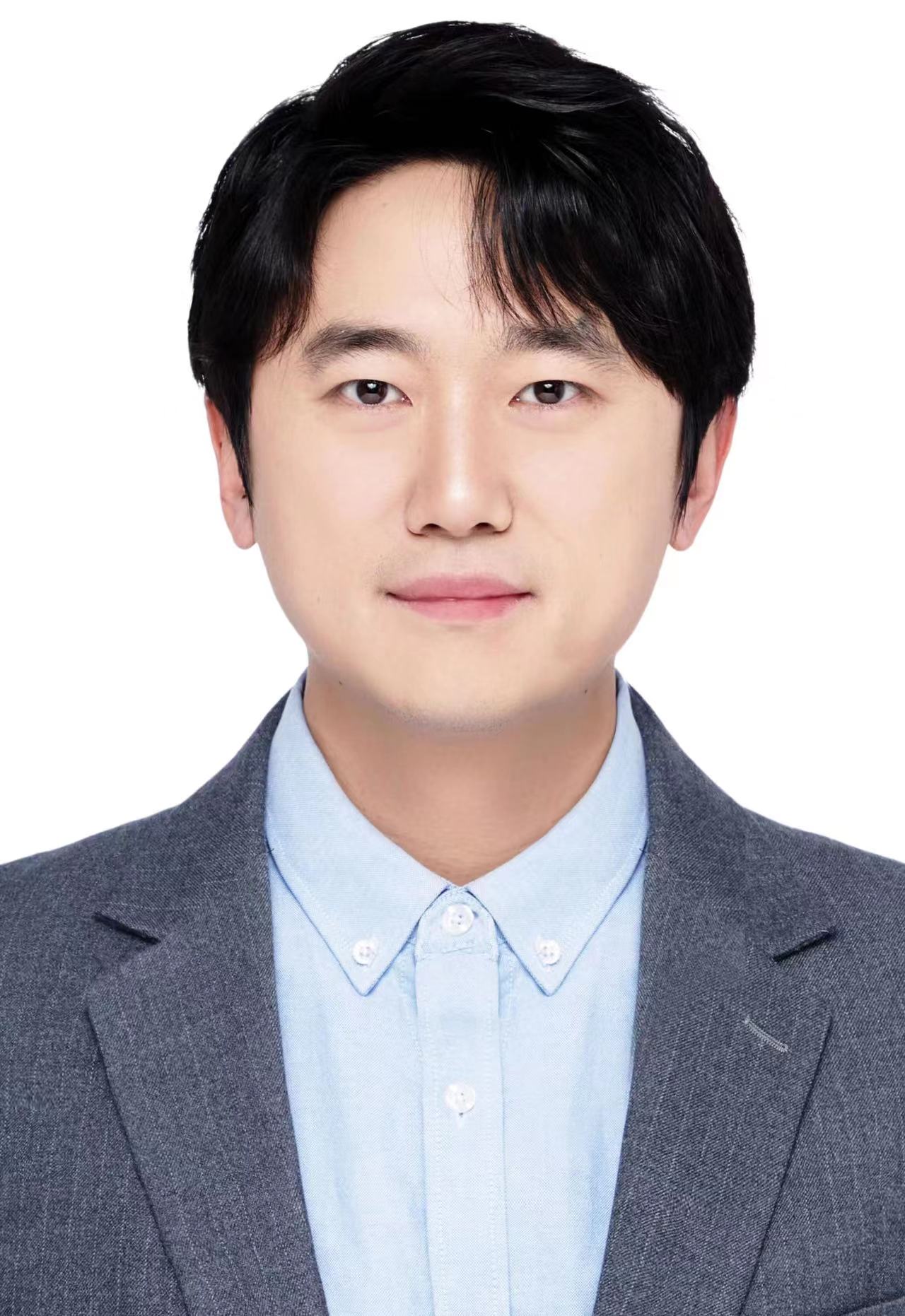}}]
	{Long Chen}
	is currently a Professor with State Key Laboratory of Management and Control for Complex Systems, Institute of Automation, Chinese Academy of Sciences, Beijing, China. His research interests include autonomous driving, robotics, and artificial intelligence, where he has contributed more than 100 publications. He received the IEEE Vehicular Technology Society 2018 Best Land Transportation Paper Award, the IEEE Intelligent Vehicle Symposium 2018 Best Student Paper Award and Best Workshop Paper Award, the IEEE Intelligent Transportation Systems Society 2021 Outstanding Application Award, the IEEE Conference on Digital Twin and Parallel Intelligence 2021 Best Paper and Outstanding Paper Award, the IEEE International Conference on Intelligent Transportation Systems 2021 Best Paper Award. He serves as an Associate Editor for the IEEE Transaction on Intelligent Transportation Systems, the IEEE/CAA Journal of Automatica Sinica, the IEEE Transaction on Intelligent Vehicle and the IEEE Technical Committee on Cyber-Physical Systems.

\end{IEEEbiography}

\end{document}